\documentclass[conference,10pt]{IEEEtran}

\usepackage{framed}
\usepackage{algorithm}
\usepackage[noend]{algpseudocode}
\usepackage{bm}
\usepackage{stackrel}
\usepackage{subfigure}
\usepackage{epsf}
\usepackage{amsmath,amssymb}
\usepackage{graphicx}
\usepackage{color}
\usepackage{cite}
\usepackage{multirow,tabularx}
\usepackage{ifthen}
\usepackage{epstopdf}

\input{epstopdf.sty}
\usepackage{times}

\input{epsf.sty}

\makeatletter
\def\BState{\State\hskip-\ALG@thistlm}
\makeatother

\newcommand{\hide}[1]{\ifthenelse{\boolean{false}}{#1}{}}

\newtheorem{theorem}{{\bf Theorem}}

\newtheorem{lemma}{{\bf Lemma}}

\newtheorem{definition}{{\bf Definition}}
\newtheorem{assumption}{{\bf Assumption}}

\newcommand{\qed}{\nobreak \ifvmode \relax \else
      \ifdim\lastskip<1.5em \hskip-\lastskip
      \hskip1.5em plus0em minus0.5em \fi \nobreak
      \vrule height0.75em width0.5em depth0.25em\fi}

\newcommand{\beq}{\begin{equation}}
\newcommand{\eeq}{\end{equation}}
\newcommand{\barr}{\begin{array}}
\newcommand{\earr}{\end{array}}

\newcommand{\benum}{\begin{enumerate}}
\newcommand{\eenum}{\end{enumerate}}

\newcommand{\bit}{\begin{itemize}}
\newcommand{\eit}{\end{itemize}}

\newcommand{\bc}{\begin{center}}
\newcommand{\ec}{\end{center}}

\newcommand{\bdes}{\begin{description}}
\newcommand{\edes}{\end{description}}

\newcommand{\bfig}{\begin{figure}}
\newcommand{\efig}{\end{figure}}

\newcommand{\bemq}{\begin{quote} \begin{em}}
\newcommand{\eemq}{\end{em} \end{quote}}

\newcommand{\bmp}{\begin{minipage}}
\newcommand{\emp}{\end{minipage}}

\newcommand{\bsp}{\begin{slide*}}
\newcommand{\esp}{\end{slide*}}
\newcommand{\bsl}{\begin{slide}}
\newcommand{\esl}{\end{slide}}

\newcommand{\blem}{\begin{lemma}}
\newcommand{\elem}{\end{lemma}}
\newcommand{\bthm}{\begin{theorem}}
\newcommand{\ethm}{\end{theorem}}

\newcommand{\Rn}{\mathbb{R}^n}

\newcommand\independent{\protect\mathpalette{\protect\independenT}{\perp}}
\def\independenT#1#2{\mathrel{\rlap{$#1#2$}\mkern2mu{#1#2}}}

\newcommand{\pa}[1]{\text{Pa}_{#1}}
\newcommand{\NonDes}[1]{\text{NonDes}_{#1}}
\newcommand{\an}[1]{\text{an}(#1)}

\IEEEoverridecommandlockouts

\begin{document}



\title{A Theory of Uncertainty Variables for State Estimation and Inference}


\author{Rajat Talak, Sertac Karaman, and Eytan Modiano
\thanks{This work was supported by NSF Grants AST-1547331, CNS-1713725, and
CNS-1701964, and by Army Research Office (ARO) grant number W911NF-
17-1-0508. This paper was presented in part at the Allerton Conference in 2019~\cite{talak19_Allerton_UV_BayNet}. }
\thanks{The authors are with the Laboratory for Information and Decision Systems (LIDS) at the Massachusetts Institute of Technology (MIT), Cambridge, MA. {\tt \{talak, sertac, modiano\}@mit.edu}}
}

\IEEEaftertitletext{\vspace{-0.6\baselineskip}}

\maketitle

\begin{abstract}
We develop a new framework of uncertainty variables to model uncertainty. An uncertainty variable is characterized by an uncertainty set, in which its realization is bound to lie, while the conditional uncertainty is characterized by a set map, from a given realization of a variable to a set of possible realizations of another variable. We prove Bayes' law and the law of total probability equivalents for uncertainty variables. We define a notion of independence, conditional independence, and pairwise independence for a collection of uncertainty variables, and show that this new notion of independence preserves the properties of independence defined over random variables. We then develop a graphical model, namely Bayesian uncertainty network, a Bayesian network equivalent defined over a collection of uncertainty variables, and show that all the natural conditional independence properties, expected out of a Bayesian network, hold for the Bayesian uncertainty network. We also define the notion of point estimate, and show its relation with the maximum a posteriori estimate.
Probability theory starts with a distribution function (equivalently a probability measure) as a primitive and builds all other useful concepts, such as law of total probability, Bayes' law, independence,  graphical models, point estimate, on it. Our work shows that it is perfectly possible to start with a set, instead of a distribution function, and retain all the useful ideas needed for state estimation and inference.
\end{abstract}

\section{Introduction}
\label{sec:intro}

Probability theory, developed over the last three centuries, has provided an overarching framework for modeling uncertainty in the real-world. As a result, it has become a key mathematical tool used in designing state estimation and inference algorithms. Pierre-Simon Laplace and Thomas Bayes were among the first to formulate the notion of conditional probability, and use it to estimate an unknown parameter from observed data~\cite{Hald_2007_ProbHistory1713Till1935, Stigler_1986_StatHistoryTill1900}. Ever since the axiomatic foundations laid by Kolmogorov~\cite{Kolmogorov_1933_ProbTheory} and the appearance of de Finetti's theorem~\cite{1978_Kingman_Exchangeability, 1985_Aldous_Exchangeability}, the theory of probability has justifiably formed the basis for inference and state estimation algorithms.

In Bayesian inference, for example, the goal is to successively improve an estimate of a model parameter or an evolving state variable, such as the pose of a robot~\cite{thurn_prob_robotics}, by incorporating the observed information~\cite{christian_robert_ml, bishop_ml}. A prior probability distribution is assigned to the initial state variable or the model parameter, and this distribution is successively improved by computing the posteriori distribution, using the observed data. Bayes' law and the law of total probability form the theoretical basis for this computation.

One of the main difficulties in such state estimation and inference procedures is its computational tractability. Computing the posteriori distribution and the maximum a posteriori (MAP) estimate is hard in most problems of practical significance~\cite{2004_Complexity_MAP_BayNet, pmlr-v99-tosh19a, bishop_ml}. Several approximation methods have been considered to overcome this limitation~\cite{2001_minka_EP, 2002_particle_filtering, Andrieu2003_mcmc, Blei2017_variational_inference, bishop_ml}, and it remains an active field of research. Graphical models such as Bayesian network leverage the underlying conditional independence structure for better inference algorithms~\cite{koller_ProbGraphModels, jordan_GraphModels}.

Another major issue with using distribution functions is that they are chosen mostly to ensure easier analysis and algorithm design. In robotic perception, for example, an additive Gaussian noise is often assumed in the motion and sensing model~\cite{thurn_prob_robotics}. Although, this produces the elegant Kalman filter solution, it can cause severe degradation in performance due to the inherent non-linearities in motion and sensing~\cite{Barefoot_2017_StateEstimation_Robotics}. In several such applications, and in robotic perception in particular, a bounded noise model may be more suited.

Probability theory, characterizes an uncertain quantity by a distribution function (or equivalently a measure function), which assigns a number to every (almost every) possible outcome. Perhaps, this distribution function is too much information to carry for computation, and results in computational intractability. The difficulty in computing the posteriori distribution is a manifestation of such intractability. Secondly, in the case of bounded, but unknown uncertainty, it may be more useful to model an uncertain quantity as a set.

Probability theory uses a distribution function as a primitive and builds all useful ideas such as the law of total probability, Bayes' law, independence, graphical models, point estimate on top of it. Is it then possible to construct an alternative theory, which replaces the distributions functions, with something much elementary, such as a set? Can we develop similar useful ideas such as law of total probability, Bayes' law, independence, graphical models, and point estimates on such a construct? Can such a theory provide a hope for devising better algorithms? In our quest towards answering these questions we develop a new theory of uncertainty variables.
\begin{figure*}
  \centering
  \includegraphics[width=0.85\linewidth]{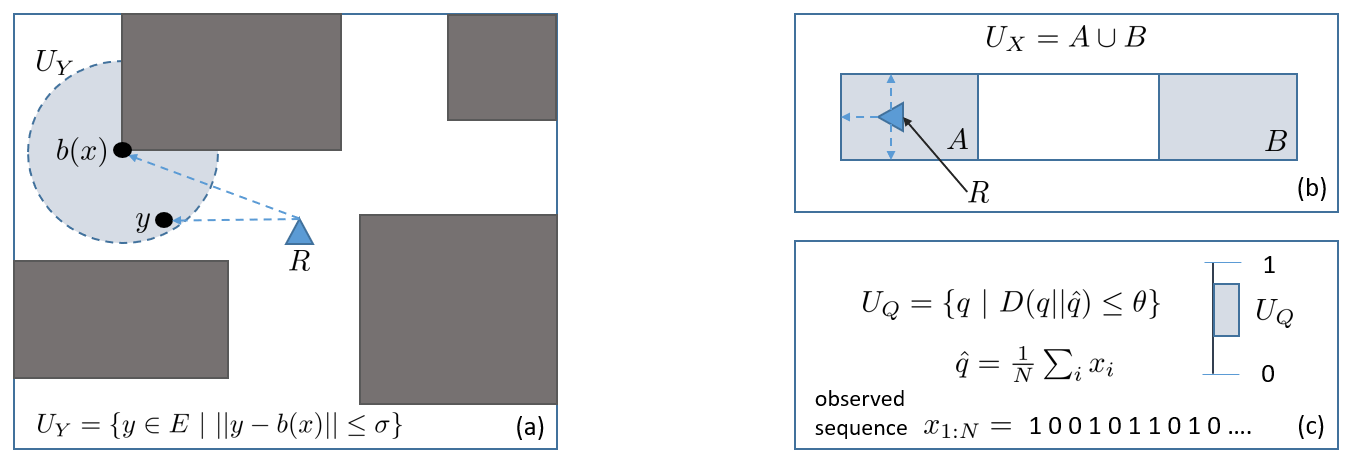}
  \caption{\textbf{(a)}~Robot $R$ is taking distance measurements of obstacle corners. Here, $x$ denotes the state space, which includes the pose of the robot as well as the configuration of obstacles (shaded regions). $b(x)$ denotes the distance of a particular corner from $R$ and $y$ its measurement. Given $x$ and $b(x)$, the measurement $y$ can be modeled as a UV $Y = (D_Y, U_Y)$ such that $D_Y = E$ and $U_Y = \{ y \in E~|~ ||y - b(x)|| \leq \sigma~\}$, where $E$ denotes the empty space (the unshaded region).
  ~\textbf{(b)}~Robot $R$ is at the corner of a tunnel and makes three distance measurements shown by dotted lines. With these measurements it can infer that its location is either in region $A$ or $B$. This is modeled by UV $X = (D_X, U_X)$, where $U_X = A\cup B$ and $D_X$ is the entire tunnel region.
  ~\textbf{(c)}~We observe a sequence of coin tosses $x_{1:N}$. $H$ and $T$ is represented by $1$ and $0$, respectively. The probability that a coin turns out $H$, namely, $q$ is more likely to lie in certain region of $[0, 1]$, given the observations $x_{1:N}$. This is modeled by a UV $Q = (D_Q, U_Q)$, where $D_Q = [0, 1]$ and $U_Q$ is shown in (c), where $D(\cdot||\cdot)$ is KL divergence.}
  \label{fig:uv_examples}
\end{figure*}

\subsection{Contribution}
Suppose we want to characterize uncertain quantities such as the measured temperature in a room, the position of a robot, noisy sensor measurements, or the state of a control system. Such uncertain quantities have an implicitly defined underlying domain. For example, a temperature measurement can take any real values, a pose of a robot is a point in a $d$ dimensional configuration space. All such  uncertain quantities are more likely to lie in certain region of this domain, and not spread out everywhere.

An uncertainty variable $X$ is characterized by a tuple $(D_X, U_X)$ where $D_X$ denotes the domain set and $U_X$ the \emph{uncertainty set} of $X$. A realization $x$ of an uncertainty variable $X$, which we write as $X = x$, can lie only in its uncertainty set, i.e. $x \in U_X$. See Figure~\ref{fig:uv_examples} for examples. Conditional uncertainty is characterized by a \emph{conditional uncertainty map} $P_{Y|X}: x \rightarrow P_{Y|X}(x) \subset D_Y$, that maps every realization $x \in U_X$ of $X$ to a subset of $D_Y$, which is the set of all realizations of $Y$, i.e. given a realization $X = x$, a realization of $Y$ can only lie in the set $P_{Y|X}(x)$. Thus, the larger the set $P_{Y|X}(x) \subset D_Y$, the larger is the uncertainty in $Y$ given $X = x$.

Using this notion of uncertainty variables and conditional uncertainty maps, we first prove the Bayes' law and the law of total probability equivalent for uncertainty variables.
We then define the notions of independence, conditional independence, and pairwise independence for a collection of uncertainty variables. We argue that this new notion of independence over uncertainty variables preserves the same properties of independence that we know from random variables. For example, we show that total independence between a collection of uncertainty variables does not imply pairwise independence.

Graphical models over random variables have been very useful in designing exact and approximate inference algorithms~\cite{koller_ProbGraphModels, jordan_GraphModels}. We extend the theory of uncertainty variables, developed in the first part of the paper, and define a graphical model over uncertainty variables.
We define \emph{Bayesian uncertainty network}, as a directed graphical model over a collection of uncertainty variables. As the name suggests, this is equivalent to the Bayesian network defined over random variables. We show that all the conditional independence properties, expected out of a Bayesian network, also hold for the Bayesian uncertainty network.

In many state estimation and inference problems, one is interested in a point estimate. We, therefore, define the notion of a point estimate. We prove a relation between the point estimate and the MAP estimate when the uncertainty sets are high-probability sets with respect to the appropriate distribution functions. 
This  illustrates the generality of this new approach of characterizing uncertainty.

\subsection{Related Works}
Using bounded sets instead of probability distributions is not a new idea, and has been explored in the control systems literature~\cite{2015_Franco_SetEstiAndControl, 1971_Ber_Thesis, 1971_BerRhodes_ReachabilityTubes}. Some of these early works on bounded noise models in control theory, also inspired the formulation of set-estimation in the signal processing literature~\cite{1989_deller_set_mem_identification_dsp, 1993_combettes_foundations_set_estimation, 2015_Franco_SetEstiAndControl}. The motivation here was that the point estimate, such as MAP or ML, is not good enough, and a confidence region, namely a set, would be useful. A set estimate, for say a model parameter, was defined as an intersection of sets, each of which corresponds to an observation. To help compute such an intersection, especially of ellipsoidal sets, several approximating methods were proposed~\cite{2002_Ros_EllipsoidalCalculus, 1997_Book_Ellipsoidal_Calculus}.


A notion of uncertainty sets has also been used in the robust optimization literature~\cite{2011_DBerm_RO_theory_n_applications, BenTal_RO_Book}. Robust optimization also begins with the same premise as ours, that the way probability theory characterizes uncertainty results in computational intractability. As a recourse, when many uncertain quantities are involved, robust optimization constructs uncertainty sets over these uncertain quantities, using the law of large numbers and the central limit theorems~\cite{2011_DBerm_RO_high_dim}. The objective is then to solve a worst case optimization problem, which is either min-max or max-min, over these uncertainty sets.

Our work, on the other hand, uses an uncertainty set instead of a distribution function, and develops a theory in parallel to the theory of probability used in state estimation and Bayesian inference. The notion of conditional uncertainty maps, independence, conditional independence, graphical models, and point estimates, developed here is novel, and does not exist in either the robust optimization or the set-estimation literature.

In~\cite{laValle_2012_sensing_filtering}, a sensor was abstractly modeled as a mapping $h$ from state space to the observation space. The preimage of the sensor mapping $h^{-1}$, evaluated at a sensor observation, gave the set of all states that could result in the particular observation. A general triangulation principle was proposed to obtain the set of all possible states, as an intersection of all sensor preimages. A general mathematical foundation for the proposed method of filtering was suggested as an open challenge in~\cite{laValle_2012_sensing_filtering}. We believe that the theory of uncertainty variables comes close to addressing this challenge. In Section~\ref{sec:bay_nets}, we will derive the general triangulation principle.

\subsection{Notations}
\label{sec:notations}
We use the following notation. For an indexed set $A$, $y_A$ denotes the collection $\{ y_i~|~i \in A \}$. We use $1:N$ to denote the set of integers $\{1, 2, \ldots N\}$. Uncertainty variables are usually denoted by $X$, $Y$, and $Z$, while random variables are denoted by $\bar{X}$, $\bar{Y}$, and $\bar{Z}$.

For a set $A$, we use $2^A$ to denote the collection of all subsets of $A$. A set cross product is denoted by $\times$. Empty set is denoted by $\emptyset$. We also use a notion of a cross product between a set and a set function. For a set $A \subset D_X$ and a set function $B: D_X \rightarrow 2^{D_Y}$, where $D_X$ and $D_Y$ are two sets, we define the cross product
\begin{align}
A\otimes B &= \cup_{x \in A}\{x\}\times B(x), \nonumber \\
           &= \{ (x,y)~|~x \in A~\text{and}~y \in B(x) \}. \nonumber
\end{align}
As an example, if $A = \{ x \in \mathbb{R}^n~|~f(x) \leq 0\}$ and $B(x) = \{ y \in \mathbb{R}^m~|~g(x,y) \leq 0 \}$, for all $x \in A$, then $A\otimes B = \{ (x,y) \in \mathbb{R}^{n+m}~|~f(x) \leq 0,~g(x,y) \leq 0~\}$.

We use the following notion of projection. If a set $D$ is such that $D = D_A\times D_B$, then the projection operator on $D_A$ is a mapping $\Pi: 2^{D} \rightarrow 2^{D_A}$, which maps every subset in $D$ to a subset in $D_A$, such that
\begin{equation}\nonumber
\Pi(U) = \{ x_A \in D_A~|~\exists~x_B~\text{s.t.}~(x_A, x_B) \in U\},
\end{equation}
for all $U \subset D$.

\subsection{Organization}
In Section~\ref{sec:theory_uv}, we develop the notion of uncertainty variables, conditional uncertainty map, and prove the two fundamental results, namely, the law of projections and Bayes' law. In Section~\ref{sec:indep}, we define independence and conditional independence over uncertainty variables, and argue that this notion retains the independence properties over random variables. In Section~\ref{sec:bay_nets}, we define the Bayesian uncertainty network and establish all the conditional independence relations it satisfies. Point estimates are discussed in~\ref{sec:pt_estimate} and we conclude in Section~\ref{sec:conc}.

\section{Theory of Uncertainty Variables}
\label{sec:theory_uv}
We first define the notion of an uncertainty variable and the conditional uncertainty map.
\begin{framed}
\begin{definition}
\label{def:uv}
An uncertainty variable (UV) $X$ is a tuple denoted as
\begin{equation}
X = (D_X, U_X),
\end{equation}
where $D_X$ is the domain of the UV and $U_X \subset D_X$ is the uncertainty set such that every realization $x$ of $X$ is in $U_X$, i.e. $x \in U_X$.
\end{definition}
\end{framed}
We will use the notation $X = x$ to denote a realization of a UV $X$. Whenever we say $X = x$ it will be presumed that $x \in U_X$. We will use upper-case letters to denote UVs and smaller-case letters to denote its realization.

Conditional uncertainty should characterize the uncertainty on one variable, say $Y$, given a realization of another variable, say $X = x$. This can be defined as a set map.
\begin{framed}
\begin{definition}
\label{def:uv_cond_map}
Let $X = (D_X, U_X)$ and $Y = (D_Y, U_Y)$ be two uncertainty variables. The conditional uncertainty map of $Y$ given $X$ is a set function $P_{Y|X}: D_X \rightarrow 2^{D_Y}$ that maps every $x \in D_X$ to a set $P_{Y|X}(x) \subset D_Y$ such that
\begin{enumerate}
  \item $P_{Y|X}(x) \neq \emptyset$ if $x \in U_X$, and
  \item $P_{Y|X}(x) \subset U_Y$ for all $x \in U_X$.
\end{enumerate}
\end{definition}
\end{framed}
The first condition enforces that whenever $x \in U_X$, $P_{Y|X}(x)$ cannot be $\emptyset$, i.e. it induces some uncertainty on variable $Y$. Note that we do not impose any condition on $P_{Y|X}(x)$ for $x \notin U_X$. The second condition makes sure that the conditional uncertainty set, given $X = x$, cannot be larger than the marginal uncertainty set $U_Y$.

We next define the joint uncertainty variable $Z = (X, Y) = (D_Z, U_Z)$ given the marginal uncertainty variable $X$ and the conditional uncertainty map $P_{Y|X}$. We use the $\otimes$ operation between a set and a set function defined in Section~\ref{sec:notations}.
\begin{framed}
\begin{definition}
\label{def:joint_uv}
Let $X = (D_X, U_X)$ and $Y = (D_Y, U_Y)$ be two uncertainty variables. Let $P_{Y|X}$ be the conditional uncertainty map of $Y$ given $X$. Then the joint uncertainty variable $Z = (X, Y)$ is defined by the domain set $D_Z = D_X\times D_Y$ and the uncertainty set
\begin{equation}
\nonumber
U_Z = U_X\otimes P_{Y|X} = \left\{ (x,y)~\left|~\begin{array}{c}
                                                 x \in U_X,~\text{and}\\
                                                 y \in P_{Y|X}(x)
                                               \end{array} \right.\right\}.
\end{equation}
\end{definition}
\end{framed}
In order to illustrate this definition, consider $U_X = \{ x \in \mathbb{R}^n~|~f(x) \leq 0\}$ for some function $f: \mathbb{R}^n \rightarrow \mathbb{R}$. Let the conditional uncertainty map be $P_{Y|X}(x) = \{ y \in \mathbb{R}^m~|~g(x,y) \leq 0\}$ for some function $g: \mathbb{R}^{n+m} \rightarrow \mathbb{R}$. Then, the joint uncertainty set for $Z = (X, Y)$ is given by
\begin{equation}
U_Z = \{ (x,y) \in \mathbb{R}^{n+m}~|~f(x) \leq 0,~g(x,y) \leq 0\}.
\end{equation}
This implies that in the characterization of the joint uncertainty set, the two variables, namely $x$ and $y$, need to satisfy both the conditions: $f(x) \leq 0$ and $g(x, y) \leq 0$. One of which defines the marginal uncertainty set $U_X$, while the other defines the conditional uncertainty map $P_{Y|X}$.

\begin{figure}
  \centering
  \includegraphics[width=0.85\linewidth]{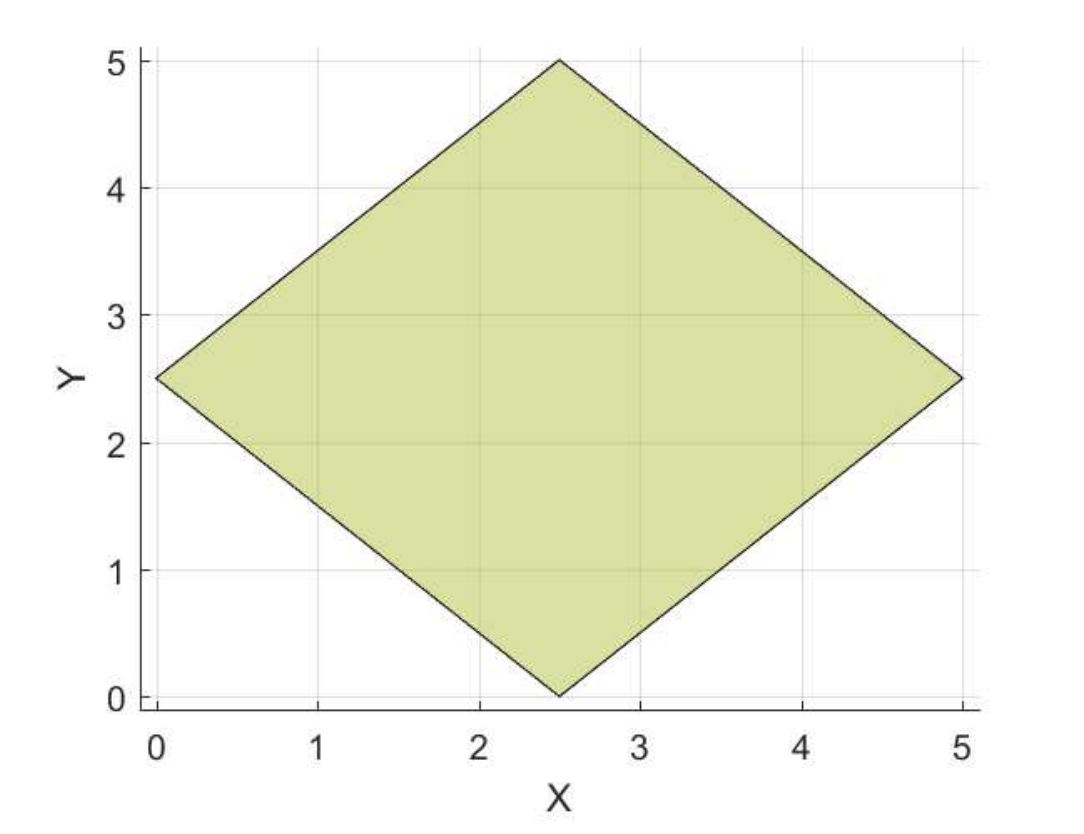}
  \caption{Plots joint uncertainity set $U_{X,Y}$ of two UVs that are not independent.}
  \label{fig:not_indep_example}
\end{figure}

It is important to note that the joint uncertainty variable $Z = (X, Y)$ is defined  with a domain that is just a cross product of the two domains $D_X \times D_Y$. However, it is not necessarily true that the joint uncertainty set $U_{X,Y}$ will also be a cross product of the marginal sets $U_X$ and $U_Y$. In Figure~\ref{fig:not_indep_example}, we provide an example. In it we have plotted the joint uncertainty set $U_{X,Y}$. Here, $U_{X} = [0, 5]$ and $P_{Y|X}$ is given by
\begin{equation}
P_{Y|X}(x) = \left\{\begin{array}{cc}
  \left[5/2 - x, 5/2 + x\right] & \text{if}~x  \in [0, 5/2]\\
  \left[x - 5/2, 15/2 - x\right] & \text{if}~x \in [5/2, 5]\\
  \emptyset &\text{otherwise}
\end{array}\right..
\end{equation}

The conditional uncertainty $P_{Y|X}$ maps each $x \in D_X$ to a set in the collection $2^{D_Y}$. The larger the set $P_{Y|X}(x)$, the greater is the uncertainty in UV $Y$, given $X = x$. For the example in Figure~\ref{fig:not_indep_example}, the conditional uncertainty in $Y$, given $X$, is the most when $X = 5/2$.

In Definition~\ref{def:uv_cond_map}, we did not impose any restriction on $P_{Y|X}(x)$ for $x \notin U_X$.
\begin{framed}
\begin{definition}
We say that the conditional uncertainty map $P_{Y|X}: D_X \rightarrow 2^{D_Y}$ is \emph{always definite} if $P_{Y|X}(x) \neq \emptyset$ for any $x \in D_X$.
\end{definition}
\end{framed}
In principle, we can set $P_{Y|X}(x)$ to any subset of $D_Y$, for $x \notin U_X$, without affecting the joint uncertainty. Therefore, assuming that the conditional uncertainty maps are always definite does not change any of the results we derive. However, it simplifies some of the proofs. Hence, without loss of generality, we make the following assumption.
\begin{framed}
\begin{assumption}
\label{ass00}
Conditional uncertainty maps are always definite.
\end{assumption}
\end{framed}

\subsection{Fundamental Laws}
\label{sec:bayes}
\label{sec:fundamental_laws}

In probability theory, the law of total probability and Bayes' law form the basis for inference and state estimation. Here, we provide equivalents of these two laws for the case of uncertainty variables.

In Definition~\ref{def:joint_uv}, we saw how the uncertainty set of a joint UV can be constructed from a marginal uncertainty set and the conditional uncertainty map. The following theorem provides a way to construct the marginal uncertainty sets and the conditional uncertainty maps, given a joint uncertainty variable.
\begin{framed}
\begin{theorem}
\label{thm:law_of_projections}
Let $Z = (D_Z, U_Z)$ be an uncertainty variable such that $D_Z = D_X\times D_Y$. Then, the two marginal uncertainty variables $X = (D_X, U_X)$ and $Y = (D_Y, U_Y)$ are such that
\begin{equation}
U_{X} = \Pi_{X}[U_{Z}]~~~\text{and}~~~U_Y = \Pi_{Y}[U_Z],
\end{equation}
where $\Pi_X$ and $\Pi_Y$ denote projection operators on $D_X$ and $D_Y$, respectively. Furthermore, the conditional uncertainty map $P_{Y|X}$ is given by
\begin{equation}
P_{Y|X}(x) = \Pi_{Y}\left[ U_{Z}\cap \{ X = x \}\right],~~~\forall~x \in U_X,
\end{equation}
where $\{X = x\}$ denotes the set $\{ (x', y') \in D_Z~|~x' = x \}$.
\end{theorem}
\end{framed}
\begin{IEEEproof}
See Appendix~\ref{pf:thm:law_of_projections}.
\end{IEEEproof}

In probability theory, the marginal distribution is obtained, from a joint distribution, by a integrating out the other variable. Theorem~\ref{thm:law_of_projections} implies that for the case of uncertainty variables the marginal uncertainty set can be obtained by a projection of the joint uncertainty set. We shall refer to this as \emph{the law of projections}. In the theory of uncertainty variables, this law is as critical as the law of total probability in probability theory.

We next prove an equivalent of Bayes' law for uncertainty variables.
\begin{framed}
\begin{theorem}
\label{thm:bayes_suv}
For the joint UV $Z = (X, Y)$, the uncertainty set is given by
\begin{equation}
\label{eq:law_bayes}
U_{X,Y} = U_X \otimes P_{Y|X} = \mathcal{R}_{YX\leftrightarrow XY}\left(U_Y \otimes P_{X|Y}\right),
\end{equation}
where $\mathcal{R}_{YX\leftrightarrow XY}\left(y, x\right) = (x, y)$ for all $x \in D_X$ and $y \in D_Y$.
\end{theorem}
\end{framed}
\begin{IEEEproof}
This result follows directly from Definition~\ref{def:joint_uv} by noting that the joint uncertainty $U_{X,Y}$ can be equivalently defined as $U_X\otimes P_{Y|X}$ or as $U_Y\otimes P_{X|Y}$, except for the change of variable ordering.
\end{IEEEproof}

In the next section, we argue that the uncertainty sets and conditional uncertainty maps, can be represented as sub-level sets of some functions. This representation will be useful in proving some of the results later in the paper. 

\subsection{Representation}
\label{sec:rep}
We have represented uncertainty variables and conditional uncertainties as sets and set functions, respectively. It is, at times, useful to deal with functions rather than sets. In this small section, we present a result, that states that every such uncertainty set or a conditional uncertainty map can be represented as a sub-level set of a function.
\begin{framed}
\begin{lemma}
\label{lem:fun_rep_suv}
The following statements are true:

\noindent 1)~An uncertainty set $U_X \subset D_X$ can be written as
        \begin{equation}
        U_X = \left\{ x \in D_X~|~ H_X(x) \leq h_X \right\},
        \end{equation}
        for some function $H_X:D_X \rightarrow \mathbb{R}^m$, $h_X \in \mathbb{R}^m$, and some positive integer $m$.

\noindent 2)~A conditional uncertainty map $P_{Y|X}:D_X \rightarrow 2^{D_Y}$ can be written as
      \begin{equation}
      P_{Y|X}(x) = \left\{ y \in D_Y~|~ H_{Y|X}(y, x) \leq h_{Y|X} \right\},
      \end{equation}
      for some function $H_{Y|X}:D_Y\times D_X \rightarrow \mathbb{R}^m$, $h_{Y|X} \in \mathbb{R}^m$, and some positive integer $m$.
\end{lemma}
\end{framed}
\begin{IEEEproof}
The proof is trivial, as such functions, namely $H_X$ and $H_{Y|X}$, can always be obtained by a simple construction. For the first part, given a set $U_X \subset D_X$, take $H_X(x) = 1 - \mathbb{I}_{U_X}(x)$, for all $x \in D_X$. Here, $\mathbb{I}_{U_X}(x)$ is the indicator function for the set $U_X$. Take $m = 1$ and $h_X = 1/2$. Then, $U_X = \{x \in D_X~|~H_X(x) \leq h_X \}$. Similarly, for the second part, take $H_{Y|X}(y,x) = 1 - \mathbb{I}_{P_{Y|X}(x)}(y)$, $m = 1$, and $h_{Y|X} = 1/2$.
\end{IEEEproof}
Note that we have not imposed any conditions on the functions $H_{X}$ and $H_{Y|X}$ in Lemma~\ref{lem:fun_rep_suv}, except that they take values in some Euclidean space $\mathbb{R}^{m}$.

In the following, we provide three parametric uncertainty variables, which may be useful in practice. These are obtained by restricting $H_X$ in Lemma~\ref{lem:fun_rep_suv} to a specific function class.

\noindent (1)~\textbf{Elliptic UV}: An Elliptic UV is defined as
\begin{equation}
\label{eq:uv_elliptic}
X = \left(\Rn, \{ x \in \Rn~|~(x-\bar{x})^{T}Q^{-1}(x-\bar{x}) \leq \eta \} \right),
\end{equation}
where $Q$ is a positive definite matrix and $\bar{x}$ is a vector in $D_X = \mathbb{R}^n$. This UV can be used to model noisy measurement of a location $\bar{x}$.

\noindent (2)~\textbf{Polytopic UV}: A polytopic UV is defined as
\begin{equation}
\label{eq:uv_polytope}
X = \left(\Rn, \{ x \in \Rn~|~H(x-\bar{x}) \leq h\}\right),
\end{equation}
where $H$ is a matrix, and $h$ and $\bar{x}$ are vectors in $\mathbb{R}^n$.

\noindent (3)~\textbf{Canonical UV}: For every random variable $\bar{X}$, taking values in $D_X$ with a probability density function $f_{\bar{X}}(x)$, we can construct a simple canonical UV $X = (D_X, U_X)$. We call it the canonical UV -- canonical to the random variable $\bar{X}$. The canonical UV $X$ is given by
\begin{equation}
\label{eq:uv_cannonical}
X = \left( D_X, \{x \in D_X~|~ -\log f_{\bar{X}}(x) \leq \eta \}\right),
\end{equation}
for some $\eta  \in \mathbb{R}$. Note that the Elliptic UV in~\eqref{eq:uv_elliptic} is a Canonical UV for the Gaussian random variable $\mathcal{N}(\bar{x}, Q)$ and the polytopic UV in~\eqref{eq:uv_polytope} is a Canonical UV for a uniformly distributed random variable over the polytope.

In the next section, we illustrate the usefulness of the uncertainty variables in computing the posteriori distribution. The main tool is the application of the law of total projections (Theorem~\ref{thm:law_of_projections}) and the Bayes' rule (Theorem~\ref{thm:bayes_suv}).

\subsection{Computing the Posteriori Map}
\label{sec:posteriori_examples}
The main advantage of this formulation is that it can be easier to compute the posteriori uncertainty map. For example, in many machine learning applications, we are given a model for the data, say $Y$, and a model for the prior parameters, say $X$. This is equivalent to knowing the conditional uncertainty map $P_{Y|X}$ and the uncertainty set $U_X$. With this, the joint uncertainty set can be computed as
\begin{equation}
U_{X,Y} = U_{X}\otimes P_{Y|X}.
\end{equation}
Then, the posteriori uncertainty map $P_{X|Y}:D_Y \rightarrow 2^{D_X}$ can be computed by a simple projection on $D_X$ (see Theorem~\ref{thm:law_of_projections}):
\begin{equation}
P_{X|Y}(y) = \Pi_X\left( U_{X,Y} \cap \{ Y = y \}\right).
\end{equation}
This posteriori map, for a given observed data $Y = y$, will produce a set in $D_X$ that tells us about the uncertainty in $X$ given the observed data $Y = y$.

Let us use the sub-level set representations of Lemma~\ref{lem:fun_rep_suv}. Let
\begin{align}
U_X &= \{ x \in D_X~|~H_{X}(x) \leq h_X\}~\text{and} \\
P_{Y|X}(x) &= \{ y \in D_Y~|~H_{Y|X}(y,x) \leq h_{Y|X}\}.
\end{align}
Then the posteriori uncertainty map $P_{X|Y}(y)$, for a given observed data $Y = y$, is given by
\begin{equation}
\label{eq:no1}
P_{X|Y}(y) = \left\{ x \in D_X~\left|~\begin{array}{c}
                                        H_{X}(x) \leq h_X~\text{and} \\
                                        H_{Y|X}(y,x) \leq h_{Y|X}
                                      \end{array}
\right.~\right\}.
\end{equation}

To see the meaning in~\eqref{eq:no1}, we define an information map $\mathcal{I}_{X|Y}$ for every conditional uncertainty map $P_{Y|X}$.
\begin{framed}
\begin{definition}
\label{def:info_map}
Information map for $P_{Y|X}$ is given by
\begin{equation}\label{eq:info_map}
\mathcal{I}_{X|Y}(y) = \{ x \in D_X~|~y \in P_{Y|X}(x)\},
\end{equation}
for every $y \in D_Y$.
\end{definition}
\end{framed}
The information map is, in a sense, an inverse of $P_{Y|X}$. It measures the set of all $x \in D_X$ which can produce an observation $y$, with the model $P_{Y|X}$. Note that for $P_{Y|X}(x) = \{ y \in D_Y~|~H_{Y|X}(y,x) \leq h_{Y|X}\}$ the information map is
\begin{equation}
\mathcal{I}_{X|Y}(y) = \{ x \in D_X~|~H_{Y|X}(y,x) \leq h_{Y|X}\}.
\end{equation}
The posteriori map in~\eqref{eq:no1} can be written as
\begin{equation}
P_{X|Y}(y) = U_X\cap \mathcal{I}_{X|Y}(y),
\end{equation}
which is the intersection of the prior uncertainty in $X$ and the uncertainty induced by the observation $Y = y$ on variable $X$, namely $\mathcal{I}_{X|Y}(y)$.

The idea of obtaining set-estimates, as intersection of sets, existed in the set-estimation literature~\cite{1989_deller_set_mem_identification_dsp, 1993_combettes_foundations_set_estimation, 2015_Franco_SetEstiAndControl}. However, the literature mostly limited itself to linear models, in which, the observed data $Y$ and the underlying state variable $X$ were related by a linear equation. Furthermore, it was not clear as to why an intersection would make sense over any other set operation. The theory of uncertainty variables developed here provides the answer. 


In the next section, we define the notion of independence and conditional independence for a given set of uncertainty variables. We show that all the independence properties that are true for random variables, such a total independence not implying pairwise independence and more, are retained for the uncertainty variables.

\section{Independence}
\label{sec:indep}
We first define independence between two uncertainty variables.
\begin{framed}
\begin{definition}
We say that the two UVs, $X$ and $Y$, are independent if $P_{Y|X}(x) = P_{Y|X}(x')$ for all $x, x' \in U_X$.
\end{definition}
\end{framed}
It is trivial to see that for independent uncertainty variables $X$ and $Y$, the joint uncertainty set also factors into the product of the marginal uncertainty set. We articulate this in the following lemma.

\begin{framed}
\begin{lemma}
Uncertainty variables $X$ and $Y$ are independent if and only if $U_{X,Y} = U_X \times U_Y$, where $U_X$, $U_Y$, and $U_{X,Y}$ are uncertainty sets for $X$, $Y$, and $(X,Y)$, respectively.
\end{lemma}
\end{framed}
\begin{IEEEproof}
We first prove the following lemma about the operation $\otimes$. 
\begin{lemma}
\label{lem:o-times-prop}
Let $A \subset D_X$ and $B:D_X \rightarrow 2^{D_Y}$. If the mapping $B$ is such that $B(x) = B(x') = \bar{B} \subset D_Y$, for all $x, x' \in A$, then $A\otimes B = A\times \bar{B}$.
\end{lemma}
\begin{IEEEproof}
Using the definition of $A\otimes B$ we have
\begin{equation}
A\otimes B = \bigcup_{x \in A} \{ x \}\times B(x) = \bigcup_{x \in A} \{ x \}\times \bar{B}, \label{eq:nooo1}
\end{equation}
where the last equality following because of the assumption $B(x) = B(x') = \bar{B} \subset D_Y$, for all $x, x' \in A$. Now, we can take the union inside the cross product in~\eqref{eq:nooo1} to get
\begin{equation}
A\otimes B = \bigcup_{x \in A} \{ x \}\times \bar{B} = \left( \bigcup_{x \in A} \{ x \} \right)\times \bar{B},
\end{equation}
which is nothing but $A\times \bar{B}$.
\end{IEEEproof}

We first prove that, if $X$ and $Y$ are independent then $U_{X,Y} = U_X\times U_Y$. Since $P_{Y|X}(x) = P_{Y|X}(x')$ for all $x, x' \in U_X$, by Lemma~\ref{lem:o-times-prop} and Theorem~\ref{thm:bayes_suv}, we have
\begin{equation}\label{eq:nooo10}
U_{X,Y} = U_X\otimes P_{Y|X} = U_X \times \bar{B},
\end{equation}
where $\bar{B} = P_{Y|X}(x)$, for an $x \in U_X$. It now suffices to show that $\bar{B} = U_Y$. Using Theorem~\ref{thm:law_of_projections}, we get $U_Y$ to be
\begin{equation}
U_Y = \Pi_{Y}\left[ U_{X,Y} \right] = \Pi_{Y}\left[ U_{X}\times \bar{B} \right] = \bar{B}.
\end{equation}

We now show that if the joint uncertainty set factorizes, i.e. $U_{X,Y} = U_X\times U_Y$, then $X$ and $Y$ are independent. We, therefore, have to show that $P_{Y|X}(x) = P_{Y|X}(x')$ for all $x, x' \in U_X$. Again, using Theorem~\ref{thm:law_of_projections}, the conditional uncertainty map is given by
\begin{align}
P_{Y|X}(x) &= \Pi_Y\left[ U_{X,Y} \bigcap \{ X = x \}\right], \\
           &= \left\{ \begin{array}{cc}
                        U_Y &~\text{if}~x \in U_X \\
                        \emptyset &~\text{otherwise}
                      \end{array}\right.,
\end{align}
for any $x \in D_X$. This implies that $P_{Y|X}(x) = P_{Y|X}(x')$ for all $x, x' \in U_X$.
\end{IEEEproof}

Conditional independence can be similarly defined. We do so in terms of factorization of the uncertainty maps.
\begin{framed}
\begin{definition}
We say that the UVs $X = (D_X, U_X)$ and $Y = (D_Y, U_Y)$ are independent, given a UV $Z = (D_Z, U_Z)$, if
\begin{equation}
P_{X,Y|Z}(z) = P_{X|Z}(z)\times P_{Y|Z}(z),
\end{equation}
for all $z \in U_Z$.
\end{definition}
\end{framed}
We will use the notation $X \independent Y$ to denote that $X$ and $Y$ are independent, and $X \independent Y | Z$ to denote that $X$ and $Y$ are conditionally independent, given $Z$.

When it comes to several uncertainty variables, the notion of independence is as tricky as it is for the random variables. Moreover, it turns out that the independence and conditional independence properties that hold for random variables also hold for uncertainty variables. In Section~\ref{sec:bay_nets}, we will introduce Bayesian network models on a collection of uncertainty variables. We will see that the set of uncertainty variables preserve the conditional independence properties, which hold for the Bayesian network defined over random variables~\cite{koller_ProbGraphModels}.

To provide a prelude, we define pairwise and total independence between a collection of uncertainty variables. In probability theory, pairwise independence does not imply total independence between a collection of random variables. The same is true for the uncertainty variables. Let us first define pairwise and total independence for the uncertainty variables.
\begin{framed}
\begin{definition}
A collection of uncertainty variables $X_{1:N}$ is said to be

\noindent 1)~pairwise independent if for each $i, j \in [N]$, $i \neq j$, we have
        \begin{equation}
        U_{X_i, X_j} = U_{X_i}\times U_{X_j},
        \end{equation}
        where $U_{X_i}$, $U_{X_j}$, and $U_{X_i, X_j}$ are uncertainty sets for $X_i$, $X_j$, and $(X_{i},X_{j})$, respectively.

\noindent 2)~totally independent if
        \begin{equation}
        U_{X_{1:N}} = \times_{i=1}^{N}U_{X_i},
        \end{equation}
        where $U_{X_{1:N}}$ and $U_{X_i}$ are the uncertainty sets of $X_{1:N}$ and $X_i$, respectively.
\end{definition}
\end{framed}

In the following lemma, we prove that pairwise independence does not implies total independence.
\begin{framed}
\begin{theorem}
\label{thm:tot_pair_indep}
If $X_{1:N}$ are totally independent then they are also pairwise independent, but the converse is not true.
\end{theorem}
\end{framed}
\begin{IEEEproof}
\begin{figure}
\centering
\includegraphics[width=0.85\linewidth]{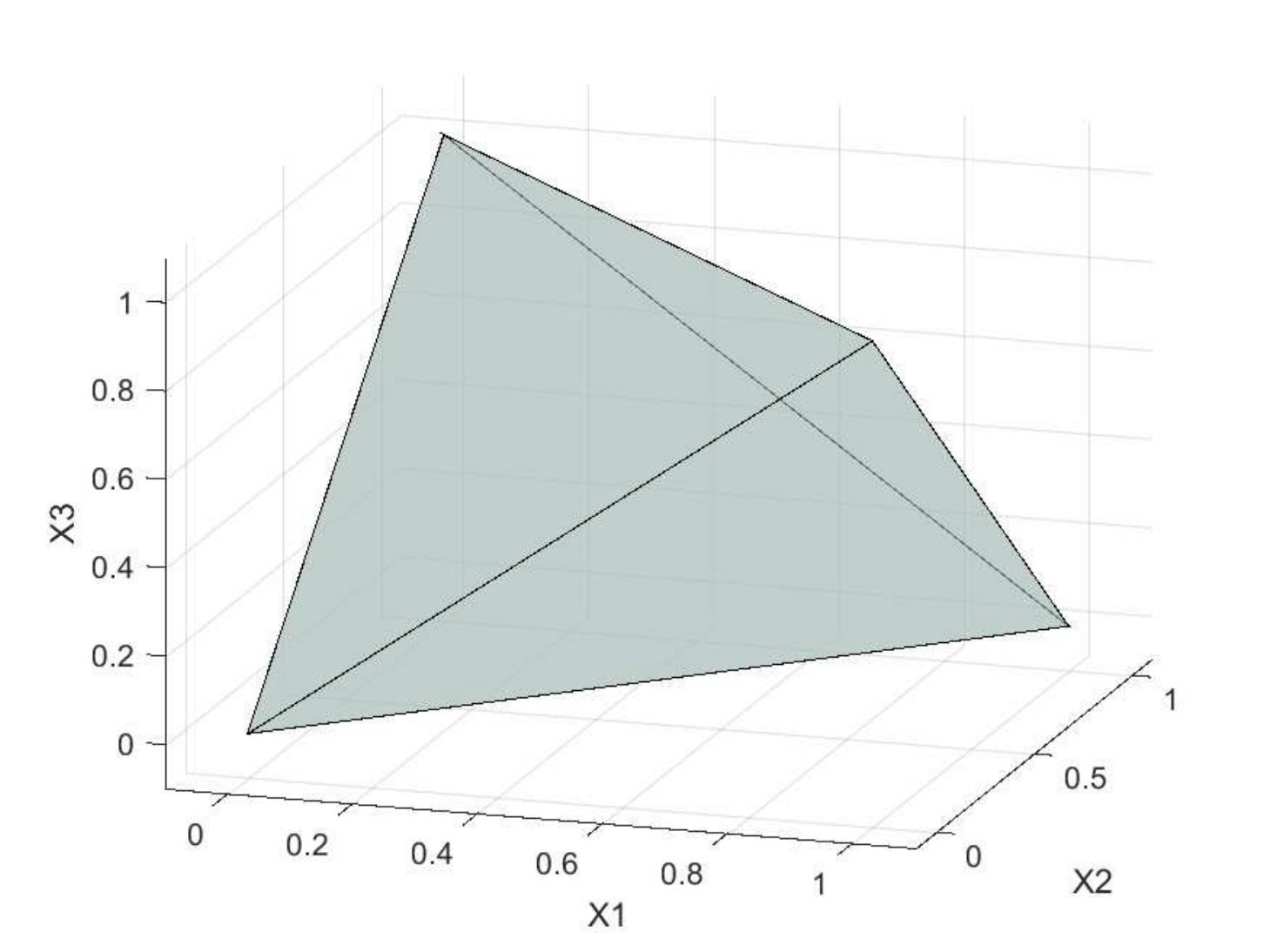}
\caption{Plot of the joint uncertainty set $U_{X_{1:3}}$ given by~\eqref{eq:independence_example}.}
\label{fig:independence_example}
\end{figure}
(a)~Let $X_{1:N}$ be totally independent uncertainty variables. Then we have $U_{X_{1:N}} = \times_{k=1}^{N}U_{X_k}$. Take $i, j \in [N]$ such that $i \neq j$. We know that the uncertainty set $U_{X_i, X_j}$ of $(X_i, X_j)$ is given by a simple projection of $U_{X_{1:N}}$ on $(X_i, X_j)$. Therefore,
\begin{align}
U_{X_i, X_j} &= \Pi_{(X_i,X_j)}\left( U_{X_{1:N}} \right), \\
            &= \Pi_{(X_i,X_j)}\left( \times_{k=1}^{N}U_{X_k} \right), \\
            &= U_{X_i}\times U_{X_j},
\end{align}
where $i$ and $j$ to be any $i, j \in [N]$ such that $i \neq j$. Thus, $X_{1:N}$ is also pairwise independent.

(b)~We prove that the converse is not true by constructing a counter-example. Take three uncertainty variables $X_{1:3}$ such that $U_{X_i} = [0, 1]$ and $U_{X_i, X_j} = [0,1]\times[0,1]$, for all $i, j \in [N]$ and $i \neq j$. However, the joint uncertainty set $U_{X_1, X_2, X_3} \neq [0,1]\times [0, 1]\times[0, 1]$. Such a joint uncertainty set $U_{X_1, X_2, X_3}$ is given by
\begin{equation}\label{eq:independence_example}
U_{X_{1:3}} = \left\{ x_{1:3}~\left|~\left[\begin{array}{rrr}
                                                1 & 1 & 1 \\
                                                -1 & -1 & 1 \\
                                                -1 &  1 & -1 \\
                                                1 & -1 & -1
                                              \end{array}\right] \left[\begin{array}{c}
                                                      x_1 \\
                                                      x_2 \\
                                                      x_3
                                                    \end{array}\right] \leq \left[\begin{array}{c}
                                                      2 \\
                                                      0 \\
                                                      0 \\
                                                      0
                                                    \end{array}\right]\right.~\right\},
\end{equation}
which is shown in Figure~\ref{fig:independence_example}.
\end{IEEEproof}

In the next section, we define the Bayesian uncertainty network, in which we extend the concept of Bayesian network, defined over a collection of random variables, to a collection of uncertainty variables. We will see that the independence properties that hold for the collection of random variables also hold for the collection of uncertainty variables. 

\section{Bayesian Uncertainty Networks}
\label{sec:bay_nets}

We now extend the notion of Bayeian network, defined for a collection of random variables, to a collection of uncertainty variables. We call it the Bayeian uncertainty network.

Let $G = (V, E)$ be a directed acyclic graph (DAG). For each node $i \in V$, let $\pa{i}$ denote the set of parents of node $i$, i.e. for each $j \in \pa{i}$ there exists a link $(j,i) \in E$. A node $k \in V$ is said to be descendant of $i$ if there exists a directed path from node $i$ to node $k$ in G. We use $\NonDes{i}$ denote the set of nodes that are non-descendants of $i$. Also, we will use $R$ to denote the set of all nodes that have no parents, i.e. $R = \{i \in V~|~\pa{i} = \emptyset \}.$ Typically, we would need to order the nodes in $V$ in a sequence.
A canonical ordering of nodes in $V$ is an ordering such that parents are indexed before their children, i.e., for all $j \in \pa{i}$, we have $j < i$. We know that such an ordering of nodes in a DAG is always possible. 


A collection of uncertainty variables is characterized by its joint uncertainty set. We now formally define the notion of \emph{Bayesian uncertainty network}, in which the uncertainty set of a collection of uncertainty variables factorizes according to an underlying DAG.
\begin{framed}
\begin{definition}
A \emph{Bayesian uncertainty network} is the tuple $\mathcal{BN} = (X_V, G)$ of uncertainty variables $X_V$ and a DAG $G = (V, E)$, such that $X_V$ \emph{factorizes} according to $G$, namely, every node $i \in V$ is associated with a unique uncertainty variable $X_i$, and
there exists conditional uncertainty maps
\begin{equation}
P_{X_i|X_{\pa{i}}}: D_{\pa{i}} \rightarrow 2^{D_i},
\end{equation}
for each $i \in V\backslash R$, such that, for any canonical ordering of nodes in $V$, the joint uncertainty set of $X_{V}$ is given by
\begin{equation}
\label{eq:uset_bun}
U_{X_{V}} = U_{X_{R}}\otimes P_{X_{|R|+1}|X_{\pa{|R|+1}}}\otimes \cdots \otimes P_{X_{|V|}|X_{\pa{|V|}}},
\end{equation}
where $U_{X_{R}}$ is a simple cross product of $U_{X_i}$, over $i \in R$, namely
\begin{equation}
U_{X_{R}} = \times_{i \in R} U_{X_i}.
\end{equation}
\end{definition}
\end{framed}

Note that the factorization in~\eqref{eq:uset_bun} is well defined, provided we ignore the ordering of variables in the tuple. To see this, let us make use of Lemma~\ref{lem:fun_rep_suv} in Section~\ref{sec:rep}. For each $i \in R$, $U_{X_i} = \{ x_i \in D_i~|~ H_{i}(x_i) \leq h_i~\}$, and for all $i \in V\backslash R$ let
\begin{equation}
P_{X_i|X_{\pa{i}}}(x_{\pa{i}}) = \{ x_i \in D_{X_i}~|~H_{i}(x_i, x_{\pa{i}}) \leq h_i~\},
\end{equation}
for some functions $H_i$ and vectors $h_i$. Then, the factorization in~\eqref{eq:uset_bun} implies that the joint uncertainty set $U_{X_{V}}$ equals
\begin{equation}\label{eq:joint_uv_baynet}
U_{X_V} = \left\{ x_V~\left|~\begin{array}{c}
                          H_{i}(x_i) \leq h_i~\forall~i \in R \\
                          H_{i}(x_i, x_{\pa{i}}) \leq h_i~\forall~i \in V\backslash R
                        \end{array}\right.\right\},
\end{equation}
This set remains the same, except for the ordering of variables in the tuple $x_{V}$. Thus, due to representation result of Lemma~\ref{lem:fun_rep_suv} we can take~\eqref{eq:joint_uv_baynet} to define the joint uncertainty set of the Bayesian uncertainty network $(X_V, G)$.

A Bayesian network, defined over random variables, satisfies many conditional independence properties. In the next section, we show that these independence properties are retained for the Baysian uncertainty network.

In Section~\ref{sec:theory_uv}, we made a simplifying assumption that the conditional uncertainty maps $P_{Y|X}$ are always definite. We argued that this does not change any of the results, but helps simplify the proofs. We make the same assumption here, and is stated as follows.
\begin{framed}
\begin{assumption}
\label{ass:0}
The conditional uncertainty map $P_{X_i|X_{\pa{i}}}$ is always definite for all $i \in V$.
\end{assumption}
\end{framed}

\subsection{Conditional Independence Properties}
We first define the local independence properties. These are a set of conditional independence properties that are satisfied by the Bayesian network. We will show that these independence properties are also valid for the Bayesian uncertainty network. 
\begin{framed}
\begin{definition}
We say that the uncertainty variables $X_{V}$ satisfy \emph{local independence properties} according to a DAG $G = (V, E)$ if

\noindent (1)~each node $i \in V$ is associated with a unique UV $X_i$, and

\noindent (2)~for every $i \in V$, we have $X_i \independent X_{\NonDes{i}} | X_{\pa{i}}$.
\end{definition}
\end{framed}


We now briefly recall the notion of d-separation in Bayesian networks. We first need to recall a few definitions. We define a path $P$ on a DAG $G$ to be a sequence of nodes $P = (i_1, i_2, \ldots i_M)$ such that either $(i_k, i_{k+1})$ or $(i_{k+1}, i_k)$ is a valid directed edge in $E$, for all $k = 1, 2, \ldots M-1$. A node $j$ on a path $P$ is said to be \emph{serial} if there exists $i, k \in P$ such that $(i, j) \in E$ and $(j, k) \in E$. Pictorially, node $j$ on path $P$ looks like $\rightarrow j \rightarrow$. Similarly, a node $j$ on path $P$ is said to be \emph{diverging} if there exists $i, k \in P$ such that $(j, i) \in E$ and $(j, k) \in E$. Pictorially, node $j$ on path $P$ looks like $\leftarrow j \rightarrow$. And finally, a node $j$ on path $P$ is said to be \emph{converging} if there exists $i, k \in P$ such that $(i, j) \in E$ and $(k, j) \in E$. Pictorially, node $j$ on path $P$ looks like $\rightarrow j \leftarrow$.

Let $A$, $B$, and $C$, be three disjoint collection of nodes in the DAG $G = (V, E)$. A path $P$ from $A$ to $B$ is a path that starts from some node in $A$ and ends at a node in $B$. We say that a path $P$ from $A$ to $B$ is \emph{blocked} by $C$ if one of the following conditions are satisfied:
\begin{enumerate}
  \item the path $P$ contains a node $j \in C$, and $j$ on $P$ is either serial or diverging
  \item the path $P$ contains a node $j \in V$, $j$ on $P$ is converging, and that $j$ and its descendants are not in $C$
\end{enumerate}

We say that $A$ and $B$ are \emph{d-separated} by $C$ if all paths from $A$ to $B$ are blocked by $C$. In the case of a Bayesian network, defined over a collection of random variables $\bar{X}_V$, it is known that if nodes $A$ and $B$ are d-separated by nodes of $C$, then the random variables $\bar{X}_A$ and $\bar{X}_B$ are independent given $\bar{X}_C$. We show that this relation of conditional independence also holds for the Bayesian uncertainty networks.

\begin{framed}
\begin{definition}
A collection of uncertainty variables $X_V$ satisfy \emph{global independence properties} with respect to a DAG $G = (V, E)$ if

\noindent (1)~each node $i \in V$ is associated with a unique uncertainty variable $X_i$, and

\noindent (2)~for all subsets $A$, $B$, and $C$ of $V$ such that $C$ d-separates $A$ and $B$ we have $X_{A}\independent X_{B} | X_{C}$.
\end{definition}
\end{framed}

We now show that the Bayesian uncertainty network satisfies the local independence property as well as the global indepdence property. Furthermore, we prove an equivalence between a collection of uncertainty variables constrained by either local independence property or global independence property and the Bayesian uncertainty network.
\begin{framed}
\begin{theorem}
\label{thm:indep_baynet}
Let $G = (V, E)$ be a DAG and $X_V$ denote a collection of uncertainty variables. The following three statements are equivalent.

\noindent (1)~$(X_V, G)$ is a Bayesian uncertainty network and Assumption~\ref{ass:0} is satisfied

\noindent (2)~$X_V$ satisfies the local independence properties with respect to $G$

\noindent (3)~$X_V$ satisfies the global independence properties with respect to $G$
\end{theorem}
\end{framed}
\begin{IEEEproof}
The fact that condition (3) implies (2) is straight forward, and can be seen by noting that $A = \{i\}$ and $B = \NonDes{i}$ are d-separated by $C = \pa{i}$ for all $i \in V$. We prove (1) implies (3) in Appendix~\ref{pf:thm:indep_baynet_1imp3} and (2) implies (1) in Appendix~\ref{pf:thm:indep_baynet_2imp1}.
\end{IEEEproof}
This theorem implies that the conditional independence properties of the Bayesian network also hold for the Bayesian uncertainty network. We discuss two simple Bayesian uncertainty networks, namely, Naive Bayes' and Kalman filtering in Appendix~\ref{app:naive_bayes} and~\ref{app:kf}, respectively. The analysis affirms the triangulation filtering principle proposed in~\cite{laValle_2012_sensing_filtering}.

In the next section, we define the notion of a point estimate. Although the idea can be generalized, we define it over a Bayesian uncertainty network. We show that if the uncertainty sets $U_{X_i}$, for $i \in R$, and the conditional uncertainty maps $P_{X_i|X_{\pa{i}}}$, for $i \in V\backslash R$, are canonical, corresponding to some distribution functions, then the defined point estimate equals to the maximum aposteriori estimate.

\section{Point Estimates}
\label{sec:pt_estimate}

In practice, we are generally interested in point estimates. For example, in the robotic estimation problem, we would like to learn the true trajectory of a robot along with the location of landmarks in its surrounding. In the regression or the classification problem, we would like to estimate the model parameters.

In this section, we define point estimate for a Bayesian uncertainty network. In the Bayesian uncertainty network, we have some uncertainty variables that we observe, and some others which we want to estimate, given the observed variables.

Let $\mathcal{BN} = (X_V, G)$ be a Bayesian uncertainty network, where $G = (V,E)$ is a DAG. Let the joint uncertainty set for $X_V$ be given by~\eqref{eq:joint_uv_baynet}. Let $J \subset V$ denote the set of nodes, which correspond to the observed data. Namely, we have $x_j = y_j$ for all $j \in J$, and that we know $y_J$. Let $I \subset V$ be the set of nodes, which correspond to the uncertainty variables that are of interest to us, and we would like to estimate. We assume $I$ and $J$ to be disjoint, and that $I \cup J = V$.

From the joint uncertainty set, we can compute the posteriori uncertainty map $P_{X_I|X_J}(x_J)$ by projection; see Theorem~\ref{thm:law_of_projections}. Evaluating $P_{X_I|X_J}(x_J)$ at the observed data $x_J = y_J$, yields a posteriori uncertainty set for $X_I$, given $X_J = y_J$. This set is given by
\begin{align}
&P_{X_I|X_J}(y_J) \\
&~~~~~= \Pi_{X_I}\left[ U_{X_V}\bigcap \{ X_J = y_J\}\right], \\
&~~~~~= \left\{ x_{I} \in D_{X_I}~\Bigg|~ \begin{array}{c}
                                                        H_{i}(x_i, x_{\pa{i}}) \leq h_i~\forall~i \in V\backslash R,  \\
                                                        H_{i}(x_i) \leq h_i~\forall~i \in R, x_J = y_J
                                                      \end{array}\right\}. \label{eq:posti_set}
\end{align}
This set gives us a sense of how uncertain we are about the variables of interest, namely $X_I$. However, it is generally required to come up with a point estimate. We define a point estimate by introducing a \emph{scaling variable} for each constraint in the posteriori set~\eqref{eq:posti_set}. These scaling variable adjust the size of each set, so as to yield an estimate. The point estimate for $X_I$, given $X_J = y_J$, is defined as
\begin{align}
\label{def:point_estimate}
\begin{aligned}
\hat{x}_I(y_{J}) =&~~\underset{x_{I},~\beta_{V}}{\text{arg min}} && \sum_{i \in V} \beta_i, \\
      &~~\text{subject to} && H_{i}(x_i, x_{\pa{i}}) \leq \beta_i h_i~\forall~i \in V\backslash R, \\
      &&& H_{i}(x_i) \leq \beta_i h_i~\forall~i \in R, \\
      &&& x_J = y_J~\text{and}~\beta_i \geq 0.
\end{aligned}
\end{align}
The optimization problem in~\eqref{def:point_estimate} is over all the variables $x_{I}$ and the scaling variables $\beta_V$. However, as the output of the argminimization, we have only shown a subset of these variables, namely $x_I$, for notational convenience.

\begin{figure}
  \centering
  \includegraphics[width=0.95\linewidth]{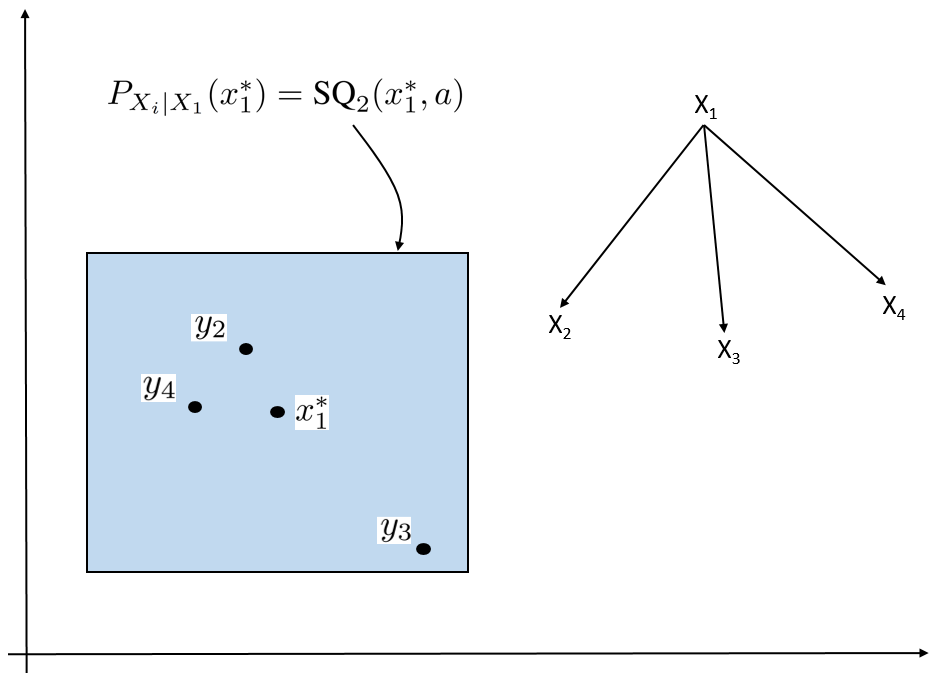}
  \caption{A Bayesian uncertainty network of four variables $X_{1:4}$. Here, $X_1 = (\mathbb{R}^2, U_{X_1} = \mathbb{R}^2)$, and the conditional uncertainty map $P_{X_i|X_1}(x^{\ast}_1) = \text{SQ}_{2}(x^{\ast}_1, a)$ is illustrated. Also, shown is the true value $x_{1}^{\ast}$ of $X_1$, and the observations $y_{2:4}$ of $X_{2:4}$.}
  \label{fig:Example_PtEstimate_Fig1}
\end{figure}

To illustrate the point estimate generated by the optimization problem~\eqref{def:point_estimate}, and the result of scaling variables $\beta_i$, we consider a simple example. Consider a Bayesian uncertainty network of four variables $X_{1:4}$ shown in Figure~\ref{fig:Example_PtEstimate_Fig1}. Here, $X_i = (\mathbb{R}^2, U_{X_i})$ for all $i$. The uncertainty set for $X_1$ is $U_{X_1} = \mathbb{R}^2$, and the conditional uncertainty maps $P_{X_i|X_1}(x_1) = \text{SQ}_{2}(x_1, a)$ for all $x_1 \in \mathbb{R}^2$, where $\text{SQ}_{2}(z, a)$ denotes a square centered at $z \in \mathbb{R}^2$ with side length $a$. The true value of the uncertainty $X_1$, namely, $x^{\ast}_1$ and the set $P_{X_i|X_1}(x^{\ast}_1) = \text{SQ}_{2}(x^{\ast}_1, a)$ is illustrated in Figure~\ref{fig:Example_PtEstimate_Fig1}.

\begin{figure}
  \centering
  \includegraphics[width=0.95\linewidth]{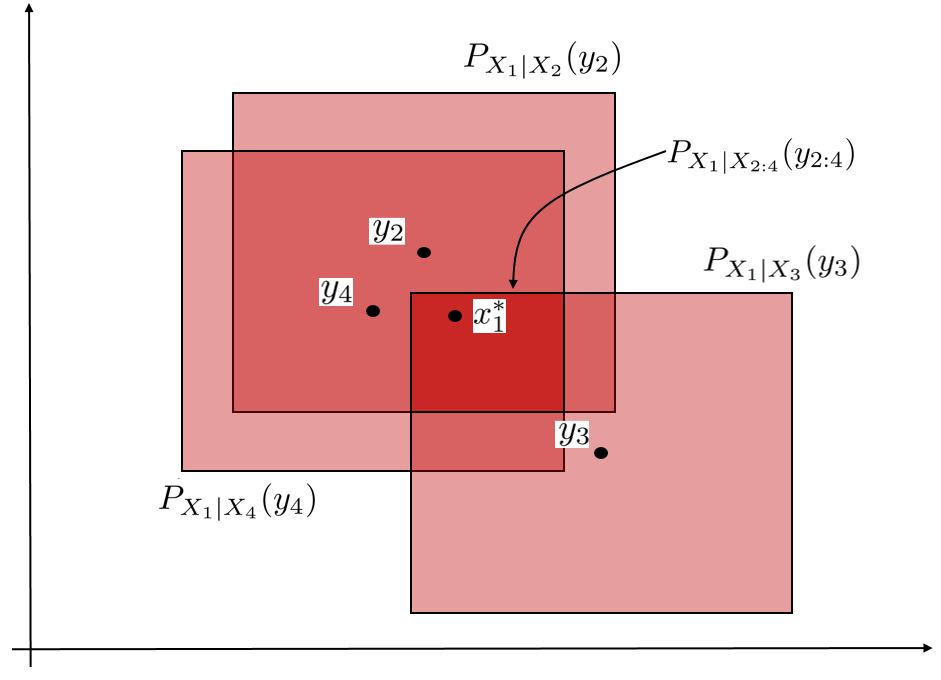}
  \caption{Illustration of the posteriori invertibility set $P_{X_1|X_{2:4}}(y_{2:4})$ as the intersection of three sets, one for each observation.}\label{fig:Example_PtEstimate_Fig3}
\end{figure}
\begin{figure}
  \centering
  \includegraphics[width=0.95\linewidth]{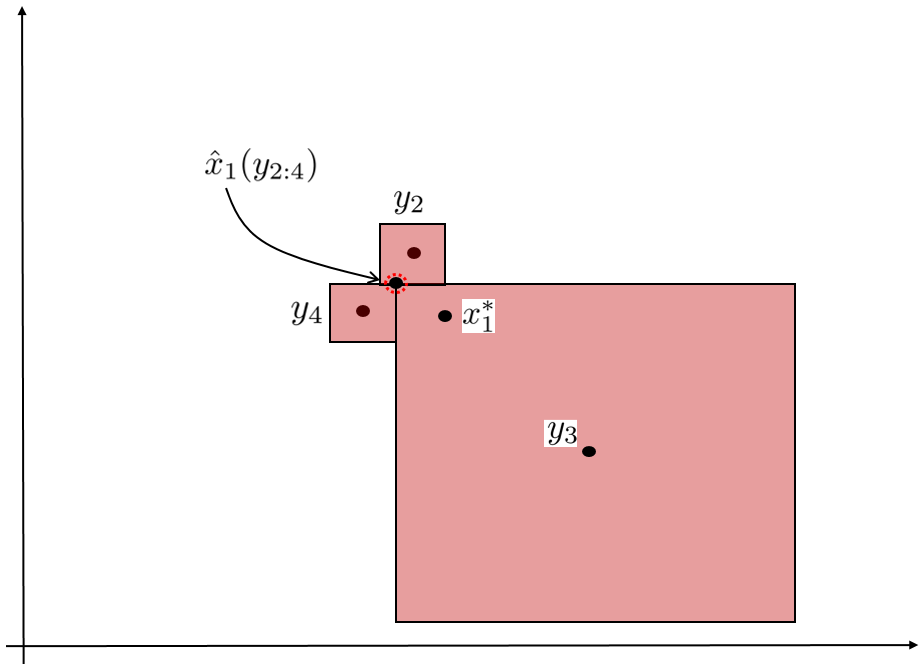}
  \caption{Shows the point estimate $\hat{x}_{1}(y_{2:4})$ at the intersection of new, minimally scaled rectangles, obtained by solving~\eqref{def:point_estimate}.}\label{fig:Example_PtEstimate_Fig4}
\end{figure}

We do not know the true value $x^{\ast}_1$ for $X_1$, and wish to estimate it by observing the variables $X_{2:4}$. Let $y_{2:4}$ be the observations of  the uncertainty variables $X_{2:4}$. Using these, we can construct a posteriori uncertainty set for $X_1$, by evaluating the posteriori uncertainty map $P_{X_1|X_{2:4}}(x_{2:4})$ at $x_{2:4} = y_{2:4}$. This gives the dark-red region shown in Figure~\ref{fig:Example_PtEstimate_Fig3}, which is the posteriori uncertainty set.

To obtain the point estimate we introduce scaling parameters $\beta_i$s, which scale the size of each of the red-colored rectangles in Figure~\ref{fig:Example_PtEstimate_Fig3}, so that they intersect only at the boundary points. The estimate $\hat{x}_{1}(y_{2:4})$ is shown in Figure~\ref{fig:Example_PtEstimate_Fig4}. We see that the rectangle corresponding to the one `far away' observation is enlarged, where as those corresponding to the other observations, that are more closer to one another, are shrunk. This is a process implicit in the definition of the point estimate~\eqref{def:point_estimate}, by which, in computing the point estimate, it weighs more in favor of observations that are closer to one another, than the one that is farther away.

Next, we show a relation between the point estimate and the MAP estimate. Before we proceed, we note that the point estimate defined in~\eqref{def:point_estimate} is not unique, and depends on the functions $H_i$ used to represent the conditional uncertainty maps $P_{X_{i}|X_{\pa{i}}}$. For example, consider the specific case in which $H_i(x_i, x_{\pa{i}}) \in \mathbb{R}$ and $h_i \in \mathbb{R}$ for all $i \in V$. Let $\Psi: \mathbb{R} \rightarrow \mathbb{R}$ be any increasing function. Then, the posteriori uncertainty set in~\eqref{eq:posti_set} can also be written as
\begin{align}
&P_{X_I|X_J}(y_J) \\
&~~= \Pi_{X_I}\left[ U_{X_V}\bigcap \{ X_J = y_J\}\right], \\
&~~= \left\{ x_{I} \in D_{X_I}\!\!~\Bigg|~\!\!\! \begin{array}{c}
                                                        \Psi(H_{i}(x_i, x_{\pa{i}})) \leq \Psi(h_i)~\forall~i \in V\backslash R,  \\
                                                        \Psi(H_{i}(x_i)) \leq \Psi(h_i)~\forall~i \in R, x_J = y_J
                                                      \end{array}\!\!\!\right\}. 
\end{align}
Thus, the point estimate will now equal
\begin{align}
\begin{aligned}
\hat{x}_I(y_{J}) =&~~\underset{x_{I},~\beta_{V}}{\text{ArgMinimize}} && \sum_{i \in V} \beta_i, \\
      &~~\text{subject to} && \Psi(H_{i}(x_i, x_{\pa{i}})) \leq \beta_i \Psi(h_i)\\
      &&&~~~~~~~~~~~~~~~~~~~~~~~\forall~i \in V\backslash R, \\
      &&& \Psi(H_{i}(x_i)) \leq \beta_i \Psi(h_i)~\forall~i \in R, \\
      &&& x_J = y_J~\text{and}~\beta_i \geq 0,
\end{aligned}
\end{align}
which is different from~\eqref{def:point_estimate}. The choice of the functions $H_i$, and $\Psi$, will have direct implication for the computational complexity of the estimate as well as the accuracy and robustness of the estimate. Several functions have been used in the literature to ensure robustness or risk-sensitivity~\cite{1981_Whittle_Risk_Sensitive_LQGC, 2019_Vasileios_OutlierRejection, 2002tac_risk_sensitive_filtering, 1997tac_risk_sensitive_filtering_smoothing}. We leave a deeper investigation into this for our future work. 

In the next section, we show a relation between the point estimate defined here, for a Bayesian uncertainty network, and the MAP estimate of a canonical Bayesian network.




\subsection{Relation with MAP}
\label{sec:map_ml}
In this section, we show a relation between the MAP and ML estimate of a Bayesian network, and the point estimate. A Bayesian network $\mathcal{B} = (\bar{X}_V, G)$ is a tuple of a collection of random variables $\bar{X}_V$ and a DAG $G = (V, E)$. For each $i \in V$, is associated a unique random variable $\bar{X}_i$ in $\bar{X}_V$. Further, for each $i \in V$, a conditional probability density\footnote{We will restrict here to the case of continuous distributions for the ease of presentation. However, these results can be extended to discrete valued random variables as well.} function $Q_{\bar{X}_i|\bar{X}_{\pa{i}}}(x_i~|~x_{\pa{i}})$ is defined. The joint density function for $\bar{X}_V$ is given by the product factorization
\begin{equation}
\label{eq:baynet_product_prob}
Q_{\bar{X}_V}(x_V) = \prod_{i \in V}Q_{\bar{X}_i|\bar{X}_{\pa{i}}}(x_i~|~x_{\pa{i}}).
\end{equation}
In what follows, we will use $Q$ to denote the probabilities.

For a given Bayesian network $\mathcal{B} = (\bar{X}_V, G)$, defined over the collection of random variables, we construct a \emph{canonical Bayesian uncertainty network} $\mathcal{BN} = (X_V, G)$, such that the underlying DAG is the same, and the functions $H_i$ and $h_i$ in~\eqref{eq:joint_uv_baynet} are given by
\begin{equation}
\label{eq:canonical_fun}
H_i(x_i, x_{\pa{i}}) = -\log \left( Q_{\bar{X}_i|\bar{X}_{\pa{i}}}(x_i~|~x_{\pa{i}}) \right),
\end{equation}
and $h_i = \eta \in \mathbb{R}$, for all $i \in V$. Note that for all $i \in R$, $\pa{i} = \emptyset$, and therefore $H_i$ reduces to a function of just $x_i$.

We now show that the point estimate for the canonical Bayesian uncertainty network, equals the MAP estimate for the corresponding Bayesian network.
\begin{framed}
\begin{theorem}
\label{thm:map}
For the canonical Bayesian uncertainty network $\mathcal{BN} = (X_V, G)$,
\begin{equation}
\hat{x}_I(y_J) = \arg\max_{x_I} Q_{\bar{X}_I|\bar{X}_J}(x_I|y_J),
\end{equation}
where $Q_{\bar{X}_I|\bar{X}_J}(x_I|y_J)$ denotes the probability density function of $\bar{X}_I$ given $\bar{X}_J$.
\end{theorem}
\end{framed}
\begin{IEEEproof}
Note that for the canonical Bayesian uncertainty network $\mathcal{BN} = (X_V, G)$, the inequality constraints in~\eqref{def:point_estimate} take the form:
\begin{equation}
-\log\left( Q_{\bar{X}_i|\bar{X}_{\pa{i}}}(x_i|x_{\pa{i}}) \right) \leq \beta_i \eta,
\end{equation}
for all $i \in V$. Furthermore, at optimality, all these constraints must be satisfied with equality. As otherwise, $\beta_i$ can be reduced to yield a smaller value of the objective function in~\eqref{def:point_estimate}. Therefore, for optimality, we have
\begin{equation}
\beta_i = -\frac{1}{\eta}\log\left( Q_{\bar{X}_i|\bar{X}_{\pa{i}}}(x_i|x_{\pa{i}}) \right),
\end{equation}
for all $i \in V$. Substituting this in~\eqref{def:point_estimate}, the optimization problem reduces to
\begin{align}
\label{pf:thm:map_eq1}
\begin{aligned}
\hat{x}_I(y_{J}) =&~~\underset{x_{I}}{\text{ArgMinimize}} \!\!\!\! && -\frac{1}{\eta}\!\sum_{i \in V} \! \log\left( Q_{\bar{X}_i|\bar{X}_{\pa{i}}}(x_i|x_{\pa{i}}) \right), \\
      &~~\text{subject to} && x_J = y_J.
\end{aligned}
\end{align}
Taking the sum inside the $\log$, as a product, and using~\eqref{eq:baynet_product_prob} we see that the objective function equals $- \frac{1}{\eta}\log\left( Q_{\bar{X}_V}(x_V)\right)$. Thus,~\eqref{pf:thm:map_eq1} equals
\begin{align}
\label{pf:thm:map_eq2}
\begin{aligned}
\hat{x}_I(y_{J}) =&~~\underset{x_{I}}{\text{ArgMaximize}} && Q_{\bar{X}_V}(x_I, x_J = y_J).
\end{aligned}
\end{align}
Since $Q_{\bar{X}_V}(x_I, x_J = y_J) = Q_{\bar{X}_I|\bar{X}_J}(x_I|y_J) Q_{\bar{X}_J}(y_J)$. Substituting this in~\eqref{pf:thm:map_eq2}, and removing $Q_{\bar{X}_J}(y_J)$ from the objective function, as it is a constant, yields the result.
\end{IEEEproof}
This result shows that the point estimate indeed equals the MAP estimate for a canonically defined Bayesian uncertainty network.


\section{Conclusion}
\label{sec:conc}
We developed a new framework of uncertainty variables to model uncertainty in the real world.
We proved Bayes' law and the law of total probability equivalents for uncertainty variables, and showed how this could be used in computing the posteriori uncertainty maps.
We defined a notion of independence, conditional independence, and pairwise independence for a given collection of uncertainty variables. We showed that this new notion of independence preserves the properties of independence defined over random variables.

In the second part, we developed a graphical model over a collection of uncertainty variables, namely the Bayesian uncertainty network. This was motivated by the Bayesian network defined over a collection of random variables. A Bayesian network satisfies certain natural conditional independence properties, derived out of the graph structure. We showed that all the natural conditional independence properties, expected out of a Bayesian network, hold also for the Bayesian uncertainty network.
We defined a notion of point estimate and showed its relation with the maximum a posteriori estimate.

In a follow up work, we will apply this theory to develop new algorithms for problems in robotic perception and planning.

\appendix

\subsection{Proof of Theorem~\ref{thm:law_of_projections}}
\label{pf:thm:law_of_projections}
The proof follows by simple application of definitions. From Definition~\ref{def:joint_uv}, we know that
\begin{equation}
U_{X,Y} = \{ (x,y) \in D_X\times D_Y~|~x \in U_X~\text{and}~y \in P_{Y|X}(x)\}.
\end{equation}
Taking projection on $D_X$ we get
\begin{equation}
\Pi_X[U_{X,Y}] = \{ x \in D_X~|~x \in U_X~\text{and}~\exists~y \in P_{Y|X}(x)\}.
\end{equation}
Since $P_{Y|X}(x) \neq \emptyset$ whenever $x \in U_X$ (see Definition~\ref{def:uv_cond_map}), the last condition always holds true and can be ignored. This yields
\begin{equation}
\Pi_X[U_{X,Y}] = \{ x \in D_X~|~x \in U_X\} = U_X.
\end{equation}
Similarly, $\Pi_Y[U_{X,Y}] = U_Y$ can be established.

We now prove that $P_{Y|X}(x) = \Pi_{Y}[U_Z\cap \{ X = x\}]$. Note that $\{ X = x\}$ denotes the set $\{ (x,y)~|~y \in D_Y \}$. Taking its intersection with $U_{X,Y}$ we obtain
\begin{equation}
U_{X,Y}\cap \{ X = x\} = \{ (x,y)~|~y \in P_{Y|X}(x)\},
\end{equation}
the projection of which, on $D_Y$, yields
\begin{equation}
\Pi_Y[U_{X,Y}\cap \{ X = x\}] = \{ y~|~y \in P_{Y|X}(x)\} = P_{Y|X}(x).
\end{equation}
This proves the result.

\subsection{Collection of Useful Results}
\label{pf:useful_results}
Conditional independence relations between uncertainty variables induce a structure on the joint uncertainty sets. Here, we prove a few of such structural results.
The first result shows that a simple conditional independence relation $X_{A}\independent X_{B} | X_C$ induces a special structure on the joint uncertainty set.
\begin{framed}
\begin{lemma}
\label{lem:cond_indep_set_factor}
Let $X_A, X_B, X_C$ denote three uncertainty variables with a joint uncertainty set
\begin{equation}
U_{X_{A, B, C}} = \{ (x_A, x_B, x_C)~|~H(x_A, x_B, x_C) \leq h\}.
\end{equation}
Then, $X_A\independent X_B | X_C$ if and only if the joint uncertainty set $U_{X_{A, B, C}}$ has the form
\begin{equation}
\label{eq:set_factor01}
U_{X_{A, B, C}} = \left\{ (x_A, x_B, x_C)~\left|~\begin{array}{c}
                                                   \Phi_1(x_A, x_C) \leq h_1, \\
                                                   \Phi_2(x_B, x_C) \leq h_2
                                                 \end{array} \right.\right\},
\end{equation}
for some functions $\Phi_1, \Phi_2$, and vectors $h_1, h_2$.
\end{lemma}
\end{framed}
\begin{IEEEproof}
We prove the \emph{only if} part in (A) and the \emph{if} part in (B). The two, put together, proves the result.

\noindent \textbf{(A)}~Let $X_A \independent X_B | X_C$ and let $P_{X_A|X_C}(x_C) = \{ x_A~|~ \Phi_1(x_A, x_C) \leq h_1\}$ and $P_{X_B|X_C}(x_C) = \{ x_B~|~ \Phi_2(x_B, x_C) \leq h_2\}$ for some $\Phi_1, \Phi_2, h_1,$ and $h_2$. Such a representation of conditional uncertainty maps is always possible due to Lemma~\ref{lem:fun_rep_suv}. Since $X_A \independent X_B | X_C$, the conditional uncertainty map $P_{X_A, X_B| X_C}$ is given by
\begin{align}
P_{X_A, X_B| X_C}(x_C) &= P_{X_A|X_C}(x_C)\times P_{X_B|X_C}(x_C), \nonumber \\
                       &= \left\{ (x_A, x_B)~\left|~\begin{array}{c}
                                                      \Phi_1(x_A, x_C) \leq h_1 \\
                                                      \Phi_2(x_B, x_C) \leq h_2
                                                    \end{array}\right.\right\}, \nonumber
\end{align}
for any choice of $x_C$, which has the same form as~\eqref{eq:set_factor01}.

\noindent \textbf{(B)}~Let the joint uncertainty set have the form given in~\eqref{eq:set_factor01}. Applying Theorem~\ref{thm:law_of_projections}, the conditional uncertainty map of $(X_A, X_B)$, given $X_C = \bar{x}_C$, can be written as
\begin{align}
&P_{X_A, X_B | X_C}(\bar{x}_C) = \Pi_{X_A, X_B}\left[ U_{X_A, X_B, X_B} \cap \{ X_C = \bar{x}_C \}\right], \nonumber \\
                        &~~~= \left\{ (x_A, x_B)~\left|~\begin{array}{c}
                                                      \Phi_1(x_A, \bar{x}_C) \leq h_1 \\
                                                      \Phi_2(x_B, \bar{x}_C) \leq h_2
                                                    \end{array}\right.\right\}, \nonumber \\
                        &~~~= \left\{ x_A~|~\Phi_1(x_A, \bar{x}_C) \leq h_1\right\}\times \left\{ x_B~|~\Phi_2(x_B, \bar{x}_C) \leq h_2\right\}, \nonumber \\
                        &~~~= P_{X_A|X_C}(\bar{x}_C)\times P_{X_B|X_C}(\bar{x}_C). \nonumber
\end{align}
Since the choice of $\bar{x}_C$ was arbitrary, it follows that $X_A\independent X_B | X_C$.
\end{IEEEproof}

We next prove two results on the structure of the marginal uncertainty sets for a Bayesian uncertainty network. Let $(X_V, G = (V,E))$ be a Bayesian uncertainty network and, without loss of generality, the uncertainty set of $X_V$ be given by
\begin{equation}\label{eq:joint_uv_baynetX}
U_{X_V} = \left\{ x_V~\left|~\begin{array}{c}
                          H_{i}(x_i) \leq h_i~\forall~i \in R \\
                          H_{i}(x_i, x_{\pa{i}}) \leq h_i~\forall~i \in V\backslash R
                        \end{array}\right.\right\},
\end{equation}
where $R$ denotes the set of root nodes in $G$. We do not assume the DAG $G$ to be connected here, i.e., a leaf node in $G$ can as well be a root node.
\begin{framed}
\begin{lemma}
\label{lem:remove_aleaf}
Consider a Bayesian uncertainty network $(X_V, G = (V, E))$ that satisfies Assumption~\ref{ass:0}. Let $j \in V$ be a leaf node in $G$ and $V' = V\backslash \{j \}$. Then, the joint uncertainty set for variables $X_{V'}$ is given by
\begin{equation}
\nonumber
U_{X_{V'}} = \left\{ x_{V'}~\left|~\begin{array}{c}
                                H_{i}(x_i) \leq h_i~\forall~i \in R\backslash \{j \}  \\
                                H_{i}(x_i, x_{\pa{i}}) \leq h_i~\forall~i \in V\backslash \left( R\cup \{j\} \right)
                              \end{array}\right.\!\!\!\!\right\},
\end{equation}
where $A\backslash B = \{i \in A~|~ i \notin B\}$.
\end{lemma}
\end{framed}
\begin{IEEEproof}
Using Theorem~\ref{thm:law_of_projections}, we obtain the marginal uncertainty set by applying a projection operator
\begin{equation}\label{eq:middle01}
U_{X_{V'}} = \Pi_{X_{V'}}\left[ U_{X_V}\right].
\end{equation}

First, consider the case when $j \in R$. If $j \in R$, then this implies that the leaf node $j$ is also a root node. This implies that the variable $x_j$ figures in only one constraint in~\eqref{eq:joint_uv_baynetX}, and that is $H_j(x_j) \leq x_j$. This constraint does not involve any other variable $x_i$ for $i \neq j$. Thus, the marginal uncertainty set $U_{X_{V'}} = \Pi_{X_{V'}}\left[ U_{X_V}\right]$ is given by
\begin{equation}
U_{X_{V'}} = \left\{ x_{V'}~\left|~\begin{array}{c}
                                H_{i}(x_i) \leq h_i~\forall~i \in R\backslash \{j \}  \\
                                H_{i}(x_i, x_{\pa{i}}) \leq h_i~\forall~i \in V\backslash R
                              \end{array}\right.\right\},
\end{equation}
which proves the result when $j \in R$.

If $j \notin R$, then the variable $x_j$ appears in exactly one constraint in~\eqref{eq:joint_uv_baynetX}. This is because of the fact that $j$ is a leaf node, and thus, cannot be a parent of any other node in the DAG $G$ This constraint is nothing but $H_{j}(x_j, x_{\pa{j}}) \leq h_j$. This constraint induces a dependence between $x_j$ and $x_{\pa{j}}$. Let
\begin{equation}
U = \left\{ x_{V'}~\!\!\left|~\!\!\!\begin{array}{c}
                                H_{i}(x_i) \leq h_i~\forall~i \in R  \\
                                H_{i}(x_i, x_{\pa{i}}) \leq h_i~\forall~i \in V\backslash \left( R\cup \{j \} \right)
                              \end{array}\right.\!\!\!\!\right\}.
\end{equation}
Then, $U_{X_{V'}} = \Pi_{X_{V'}}\left[U_{X_V}\right] $ is given by
\begin{align}
U_{X_{V'}} &= \left\{ x_{V'}~\left|~\begin{array}{c}
                                 x_{V'} \in U~\text{and}~\exists~x_j~\text{s.t.} \\
                                 H_{j}(x_j, x_{\pa{j}}) \leq h_j
                               \end{array}\right.\right\}. \label{eq:nuro1}
\end{align}
However, due to Assumption~\ref{ass:0}, there always exists a $x_j$ such that $H_{j}(x_j, x_{\pa{j}}) \leq h_j$, for all $x_{\pa{j}}$. This implies that the last inequality in~\eqref{eq:nuro1} is redundant and can be removed. This yields $U_{X_{V'}} = U$, which proves the result.
\end{IEEEproof}

Next, we generalize Lemma~\ref{lem:remove_aleaf} in order to obtain the marginal uncertainty sets $U_{X_A}$, where $A \subset V$. We consider special class of subsets $A \subset V$. Recall that an ancestral set of $A \subset V$, denoted by $\an{A}$, is the set of all nodes in $A$ and all the ancestors of $A$. We say that the set $A \subset V$ is ancestral if  $A = \an{A}$, i.e. it contains all its ancestors. The following lemma generalizes Lemma~\ref{lem:remove_aleaf} and helps us derive the marginal uncertainty sets $U_{X_A}$, for $A \subset V$, when $A$ is ancestral.
\begin{framed}
\begin{lemma}
\label{lem:remove_des}
Consider a Bayesian uncertainty network $(X_V, G = (V, E))$ that satisfies Assumption~\ref{ass:0}.
If $A \subset V$ is an ancestral set then the marginal uncertainty set of $X_A$ is given by
\begin{equation}
U_{X_A} = \left\{ x_A~\left|~\begin{array}{c}
                               H_{i}(x_i) \leq h_i~\forall~i \in R\cap A \\
                               H_{i}(x_i, x_{\pa{i}}) \leq h_i~\forall~i \in ( V\backslash R )\cap A
                             \end{array} \right.\right\} \nonumber
\end{equation}
\end{lemma}
\end{framed}
\begin{IEEEproof}
This is proved by repeated application of Lemma~\ref{lem:remove_aleaf}. First, note that $V\backslash A$ either contains a leaf node or $V = \an{A}$. If there is no leaf node in $V\backslash A$, it means $V = \an{A} = A$ as $A$ is ancestral. Thus, we need prove nothing more as $V = A$, $R \cap A = R$, $( V\backslash R )\cap A = V\backslash R$ and thus $U_{X_A} = U_{X_V}$.

If $V \backslash A$ does contain one or more leaf nodes, pick a leaf node $j \in V\backslash A$. Apply Lemma~\ref{lem:remove_aleaf} to obtain $U_{X_{V'}}$ where $V' = V\backslash \{j\}$. Keep doing this till $V'$ contains no leaf node, which is when we will have $V' = A$. At this juncture all the constraints ($H_{i}(x_i) \leq h_i$ and $H_{i}(x_i, x_{\pa{i}}) \leq h_i$) for which $i \in A$ will survive in the uncertainty set of $U_{X_{V'}}$, while the rest will be eliminated. This yields the result.
\end{IEEEproof}




\subsection{Proof (1) implies (3) in Theorem~\ref{thm:indep_baynet}}
\label{pf:thm:indep_baynet_1imp3}
Here, we prove that condition (1) implies (3) in the Theorem~\ref{thm:indep_baynet}. We make use of the results derived in Appendix~\ref{pf:useful_results}.

Assume condition (1) to be true, i.e. $(X_V, G)$ is a Bayesian uncertainty network with the joint uncertainty set of $X_V$ given by~\eqref{eq:joint_uv_baynet} and the Assumption~\ref{ass:0} holds true. We first show that the condition (3) holds under the constraint that the three sets $A$, $B$, and $C$ span $V$, i.e. $A\cup B \cup C = V$.
\begin{framed}
\begin{lemma}
\label{lem:full_ABC}
Let $A$, $B$, and $C$ be disjoint subsets of $V$ such that $A\cup B\cup C = V$. If $C$ d-separates $A$ and $B$ then $X_A \independent X_B | X_C$.
\end{lemma}
\end{framed}
\begin{IEEEproof}
Let $C_1 = \{ j \in C~|~\exists~k \in \pa{j}~\text{s.t.}~k \in A\}$. Note that for all $j \in C_1$, $\pa{j}\cap B = \emptyset$. This is because $C$ d-separates $A$ and $B$, and that there exists a $k \in \pa{j}$ which is in $A$.
Let $C_2 = C\backslash C_1$. Note that for all $i \in C_2$, we must have $\pa{i}\cap A = \emptyset$. This follows from the definition of $C_1$ and the fact that $C_2$ is its complement.
All of this implies
\begin{equation}
\label{eq:middle02}
\pa{A\cup C_1} \subset A \cup C~~~\text{and}~~~\pa{B \cup C_2} \subset B \cup C,
\end{equation}
where $\pa{M}$ denotes the set of all parent nodes of $M \subset V$.

The fact that the joint uncertainty set $U_{X_V}$ is given by~\eqref{eq:joint_uv_baynet}, along with~\eqref{eq:middle02}, implies that $U_{X_V}$ must have the form
\begin{equation}
U_{X_V} = \left\{ (x_A, x_B, x_C)~\left|~\begin{array}{c}
                                           \Phi_{1}(x_A, x_C) \leq h_1, \\
                                           \Phi_{2}(x_B, x_C) \leq h_2
                                         \end{array}\right.\right\}, \label{eq:nuro5}
\end{equation}
for some functions $\Phi_i$ and vectors $h_i$. Applying Lemma~\ref{lem:cond_indep_set_factor} to~\eqref{eq:nuro5} proves that $X_A \independent X_B | X_C$.
\end{IEEEproof}

Using Lemma~\ref{lem:full_ABC} and Lemma~\ref{lem:cond_indep_set_factor} we now show that condition (3) holds for any choice of sets $A$, $B$, and $C$, which not necessarily span $V$.

Let $A, B \subset V$ be d-separated by $C$. Define a set $\bar{A}$ to be
\begin{equation}\nonumber
\bar{A} = \left\{ i \in V~|~\{i\}~\text{is not d-separated from}~A~\text{by}~C\right\},
\end{equation}
and $\bar{B}$ to be $\bar{B} = V\backslash ( \bar{A}\cup C )$. Note that $A \subset \bar{A}$ and  $B \subset \bar{B}$, and by its very construction, $C$ d-separates $A$ and $\bar{B}$. We first prove that, not just $A$, but $\bar{A}$ is d-separated from $\bar{B}$ by $C$.
\begin{framed}
\begin{lemma}
\label{lem:d-sep-twobars}
$C$ d-separates $\bar{A}$ and $\bar{B}$.
\end{lemma}
\end{framed}
\begin{IEEEproof}
We prove this by contradiction. We know that $A$ is d-separated from $\bar{B}$ by $C$. Therefore, it suffices to argue that $\bar{A}\backslash A$ is d-separated from $\bar{B}$ by $C$.

Let this be not true, i.e., assume that there is a $j \in \bar{A}\backslash A$ that is not d-separated from $\bar{B}$ by $C$. Then, there exists a path $P$ from $j$ to a $b \in \bar{B}$ that is not blocked by $C$. Since $j \in \bar{A}$, by definition, it is not d-separated from $A$ by $C$. Which implies that there also exist a path $P'$ from an $a \in A$ to $j$ that is not blocked by $C$. As a consequence, the augmented path $(P', P)$ is a path from $a \in A$ to $b \in \bar{B}$ that is not blocked by $C$. This is a contradiction since $A$ and $\bar{B}$ are d-separated by $C$. Thus, our assumption must be incorrect, i.e. $\bar{A}\backslash A$ indeed is d-separated from $\bar{B}$ by $C$.
\end{IEEEproof}

Now, notice that $\bar{A}\cup\bar{B}\cup C = V$ and that $C$ d-separates $\bar{A}$ and $\bar{B}$. Applying Lemma~\ref{lem:full_ABC} yields that $\bar{A}\independent \bar{B} | X_C$. Using Lemma~\ref{lem:cond_indep_set_factor} we see that the joint uncertainty set of $X_V = (X_{\bar{A}}, X_{\bar{B}}, X_C)$ has the form
\begin{align}
&U_{X_V} = \left\{ (x_{\bar{A}}, x_{\bar{B}}, x_C)~\left|~\begin{array}{c}
                                                           \Phi_{1}(x_{\bar{A}}, x_C) \leq h_1, \\
                                                           \Phi_{2}(x_{\bar{B}}, x_C) \leq h_2
                                                         \end{array}\right.\right\}, \nonumber \\
&= \left\{ (x_{A}, x_{A'}, x_{B}, x_{B'}, x_C)\!\!~\left|~\!\!\!\begin{array}{c}
                                                           \Phi_{1}(x_{A}, x_{A'}, x_C) \leq h_1, \\
                                                           \Phi_{2}(x_{B}, x_{B'}, x_C) \leq h_2
                                                         \end{array}\right.\!\!\!\!\!\right\}, \nonumber 
\end{align}
for some functions $\Phi_i$ and vectors $h_i$, where $A' = \bar{A}\backslash A$ and $B' = \bar{B}\backslash B$. Using the above expression for $U_{X_V}$ and applying Theorem~\ref{thm:law_of_projections}, the conditional uncertainty map of $(X_A, X_B)$ given $X_C = \bar{x}_C$ can be computed to be
\begin{align}
&P_{X_A, X_B| X_C}(\bar{x}_C) = \Pi_{X_A, X_B}\left[ U_{X_V}\cap \{ X_C = \bar{x}_C\}\right] \nonumber \\
&~~= \left\{ (x_A, x_B)~\left|~\begin{array}{c}
                                                            \exists~x_{A'}~\text{s.t.}~\Phi_{1}(x_A, x_{A'}, \bar{x}_C) \leq h_1, \\
                                                            \exists~x_{B'}~\text{s.t.}~\Phi_{2}(x_B, x_{B'}, \bar{x}_C) \leq h_2
                                                          \end{array} \right.\right\}, \nonumber \\
&~~= \left\{ x_A~|~\exists~x_{A'}~\text{s.t.}~\Phi_{1}(x_A, x_{A'}, \bar{x}_C) \leq h_1 \right\} \nonumber \\
&~~~~~~~~~~~~\times \left\{ x_B~\left|~\exists~x_{B'}~\text{s.t.}~\Phi_{2}(x_B, x_{B'}, \bar{x}_C) \leq h_2 \right.\right\}, \nonumber \\
&~~= P_{X_A|X_C}(\bar{x}_C)\times P_{X_B|X_C}(\bar{x}_C). \label{eq:nuro3}
\end{align}
Since $\bar{x}_C$ was an arbitrary choice,~\eqref{eq:nuro3} implies that $X_A\independent X_B | X_C$.


\subsection{Proof of (2) implies (1) in Theorem~\ref{thm:indep_baynet}}
\label{pf:thm:indep_baynet_2imp1}
Here, we prove that condition (2) implies (1) in the Theorem~\ref{thm:indep_baynet}. We make use of the results derived in Appendix~\ref{pf:useful_results}.

We prove this by induction over the size of the graph $G$. Let $n = |V|$ denote the size of the graph. For $n = 1$ conditions (1) and (2) trivially hold, and therefore, $(2) \implies (1)$ is true for $n = 1$. Let $(2) \implies (1)$ be true for any collection of $n-1$ uncertainty variables. Now consider a collection of $n$ uncertainty variables $X_V$, i.e. $|V| = n$, and let condition (2) hold for $X_V$ with respect to a DAG $G = (V, E)$.

We have to prove that the uncertainty set $U_{X_V}$ of $X_V$ has the form in~\eqref{eq:joint_uv_baynet} and that every conditional uncertainty map $P_{X_i|X_{\pa{i}}}$ is always definite. We first prove that $U_{X_V}$ has the form in~\eqref{eq:joint_uv_baynet}.

Take a leaf node $j \in V$. Form a new graph $G' = (V', E')$, where $V' = V\backslash \{j\}$ and $E'$ is a set of all edges in $E$ except those which are incident on the leaf node $j$. Since $X_V$ satisfy local independence properties with respect to $G$, so must $X_{V'}$ with respect to $G'$. Since $|V'| = n-1$, by the induction hypothesis we can claim that the uncertainty set of $X_{V'}$ has the form
\begin{equation}\label{eq:net1}
U_{X_{V'}} = \left\{ x_{V'}~\left|~ \begin{array}{c}
                                      H_i(x_i) \leq h_i~\forall~\i \in R, \\
                                      H_{i}(x_i, x_{\pa{i}}) \leq h_i~\forall~i\in V'\backslash R
                                    \end{array}\right.\right\}.
\end{equation}
where $R$ denote the set of root nodes in $G'$ and $G$.\footnote{The set of root nodes $R$ is the same for both $G$ and $G'$ because the leaf node $j$ cannot be a root node in $G$. This is because $G$ is fully connected.}

Note that the local independence property also holds at the leaf node $j$, namely, $X_j \independent X_{\NonDes{j}} | X_{\pa{j}}$. Since $j$ is a leaf node, we have $\NonDes{j} = V\backslash (\{j\}\cup \pa{j})$, and therefore $X_j \independent X_{V\backslash (\{j\}\cup \pa{j})} | X_{\pa{j}}$. Using this independence relation along with Lemma~\ref{lem:cond_indep_set_factor} implies that the $U_{X_V}$ must be of the form
\begin{equation}\label{eq:net2}
U_{X_{V}} = \left\{ x_{V}~\left|~ \begin{array}{c}
                                      \Phi_{1}(x_{V'}) \leq h_1, \\
                                      \Phi_{2}(x_j, x_{\pa{j}}) \leq h_2
                                    \end{array}\right.\right\},
\end{equation}
for some functions $\Phi_i$ and vectors $h_i$. From~\eqref{eq:net2}, we can construct $U_{X_{V'}}$ as a projection of $U_{X_{V}}$ on $X_{V'}$ by using Theorem~\ref{thm:law_of_projections}. This implies,
\begin{equation}\label{eq:net3}
U_{X_{V'}} = \left\{ x_{V'}~\left|~ \begin{array}{c}
                                      \Phi_{1}(x_{V'}) \leq h_1~\text{and}~\exists~x_j~\text{s.t.} \\
                                      \Phi_{2}(x_j, x_{\pa{j}}) \leq h_2
                                    \end{array}\right.\right\}.
\end{equation}
Note that the second constraint in~\eqref{eq:net3}, which states $\exists~x_j~\text{s.t.}~\Phi_{2}(x_j, x_{\pa{j}}) \leq h_2$, is a constraint that depends on $x_{\pa{j}}$ alone. The form of $U_{X_{V'}}$ is given by~\eqref{eq:net1}, and it has no constraint that depends only on $x_{\pa{j}}$. Therefore, it must be the case that for all $x_{\pa{j}}$ there exists a $x_j$ such that $\Phi_{2}(x_j, x_{\pa{j}}) \leq h_2$, and the constraint $\Phi_{1}(x_{V'}) \leq h_1$ in~\eqref{eq:net3} has the same form as
\begin{equation}
\left\{ H_i(x_i) \leq h_i~\forall~\i \in R~\text{and}~H_{i}(x_i, x_{\pa{i}}) \leq h_i~\forall~i\in V'\backslash R \right\}.
\end{equation}
Substituting this back in~\eqref{eq:net2} yields
\begin{equation}\label{eq:net4}
U_{X_{V}} = \left\{ x_{V}~\left|~ \begin{array}{c}
                                      H_i(x_i) \leq h_i~\forall~\i \in R, \\
                                      H_{i}(x_i, x_{\pa{i}}) \leq h_i~\forall~i\in V'\backslash R, \\
                                      \Phi_{2}(x_j, x_{\pa{j}}) \leq h_2
                                    \end{array}\right.\right\},
\end{equation}
which is of the required from in~\eqref{eq:joint_uv_baynet}.

It now suffices to prove that for $X_V$, all the conditional uncertainty maps $P_{X_i|X_{\pa{i}}}$ are always definite. Note that this holds true for all $i \neq j$ by the induction hypothesis, and it only remains to show that $P_{X_j|X_{\pa{j}}}$ is always definite. Since $x_j$ is contained only in the last constraint in~\eqref{eq:net4}, we have $P_{X_j|X_{\pa{j}}}(x_{\pa{j}}) = \{ x_j~|~\Phi_{2}(x_j, x_{\pa{j}}) \leq h_2\}$. We have already proved that for all $x_{\pa{j}}$ there exists a $x_j$ such that $\Phi_{2}(x_j, x_{\pa{j}}) \leq h_2$. This implies that $P_{X_j|X_{\pa{j}}}$ is always definite.

This proves that $(X_V, G)$ is a Bayesian uncertainty network, and that Assumption~\ref{ass:0} is satisfied for it. The choice of DAG $G$ and the collection of uncertainty variables $X_V$ was arbitrary, except that $|V| = n$. Thus, $(2) \implies (1)$  holds true for any graph of size $n$, and by the principle of mathematical induction $(2) \implies (1)$  holds true.

\subsection{Naive Bayes}
\label{app:naive_bayes}
Let $X$ be the state of a system, and $Y_{1:N}$ denote $N$ independent observations. Each observation is independent of the other given the system state. This is called the Naive Bayes model~\cite{bishop_ml}, and is a simple example of a Bayesian uncertainty network.

Given all the observations, i.e. $Y_{1:N} = y_{1:N}$, we wish to compute a posteriori uncertainty in $X$. Let
\begin{equation}
\label{eq:nb_p}
U_{X} = \{ x \in D_X~|~H_{0}(x) \leq h_{0}\},
\end{equation}
denote the prior uncertainty set of $X$ and
\begin{equation}
P_{Y_i|X}(x) = \{ y_i \in D_{Y_i}~|~H_{i}(y_i, x) \leq h_i\},
\end{equation}
denote the conditional uncertainty map for the observation $Y_i$, given $X$. Since $Y_i$s are independent given $X$ we can write the conditional uncertainty map of $Y_{1:N}$ given $X$ to be
\begin{align}
P_{Y_{1:N}|X}(x) &= P_{Y_1|X}(x)\times P_{Y_2|X}(x)\times \cdots P_{Y_N|X}(x), \\
                 &= \left\{ y_{1:N}~\left|~H_{i}(y_i, x) \leq h_i~\forall~i \in [N]~\right.\right\}.\label{eq:nb_lh}
\end{align}

Using the law of projections (Theorem~\ref{thm:law_of_projections}), along with~\eqref{eq:nb_p} and~\eqref{eq:nb_lh}, we can write the posteriori uncertainty map to be
\begin{equation}
\label{eq:naive_bayes_posti}
P_{X|Y_{1:N}}(y_{1:N}) = \left\{ x~\left| \begin{array}{c}
                                            H_{0}(x) \leq h_0~\text{and} \\
                                            H_{i}(y_i, x) \leq h_i~\forall~i \in [N]
                                          \end{array} \right.\right\}.
\end{equation}
Note that the information map for $P_{Y_i|X}$ is given by (see Definition~\ref{def:info_map})
\begin{align}
\mathcal{I}_{X|Y_i}(y_i) &= \{ x \in D_X~|~y_i \in P_{Y_i|X}(x) \}, \\
                         &= \{ x \in D_X~|~H_{i}(y_i, x) \leq h_i \}.
\end{align}
The posteriori uncertainty map in~\eqref{eq:naive_bayes_posti} can thus be written as
\begin{equation}
\nonumber
P_{X|Y_{1:N}}(y_{1:N}) = U_X \cap \mathcal{I}_{X|Y_1}(y_1) \cap \mathcal{I}_{X|Y_2}(y_2) \cdots \cap \mathcal{I}_{X|Y_N}(y_N),
\end{equation}
which is nothing but the intersection of information maps $\mathcal{I}_{X|Y_i}(y_i)$ and the prior uncertainty $U_X$. This proves the general triangulation principle proposed by Steven LaValle in~\cite{laValle_2012_sensing_filtering} and also the central axiom in the set estimation literature~\cite{1989_deller_set_mem_identification_dsp, 1993_combettes_foundations_set_estimation, 2015_Franco_SetEstiAndControl}.

\subsection{Kalman Filtering}
\label{app:kf}
A state $X_t$ evolves over time. At each time, we make observations $Y_t$ about the state $X_t$. We assume that the future state (e.g. at time $t+1$) is independent of the previous states given the current state $X_t$. Also, we assume the observation at time $t$ to depend only on $X_t$. This is the Kalman filtering model, albeit generalized.

Let $U_{X_0} = \{ x_0~|~H_{0}(x_0) \leq h_0 \}$ denote the prior uncertainty on $X_0$. Let the conditional uncertainty map for $X_t$, given $X_{t-1}$, be given by
\begin{equation}
P_{X_t|X_{t-1}}(x_{t-1}) = \{ x_t~|~H_{t}(x_t, x_{t-1}) \leq h_t \},
\end{equation}
and the conditional uncertainty map for observation $Y_t$, given $X_t$, be
\begin{equation}
P_{Y_{t}|X_{t}}(x_t) = \{ y_t~|~G_{t}(y_t, x_t) \leq g_t \},
\end{equation}
for all $t = 1, 2, \ldots T$. Using the conditional independence properties, the joint uncertainty set on $(X_{0:T},Y_{1:T})$ can be constructed to be
\begin{equation}
U_{X_{0:T}, Y_{1:T}} = \left\{ (x_{0:T}, y_{1:T}) \left| \begin{array}{c}
                                                   H_{0}(x_0) \leq h_0, \\
                                                   H_{t}(x_t, x_{t-1}) \leq h_t~\forall t \in [T], \\
                                                   G_{t}(y_t, x_t) \leq g_t~\forall t \in [T]
                                                 \end{array} \right.\right\}. \nonumber
\end{equation}
The posteriori uncertainty map, given $Y_{1:T} = y_{1:T}$, is given by
\begin{equation}
\label{eq:kf_posti}
P_{X_{0:T}|Y_{1:T}}(y_{1:T}) = \left\{ x_{0:T}\! \left|\! \begin{array}{c}
                                                   H_{0}(x_0) \leq h_0, \\
                                                   H_{t}(x_t, x_{t-1}) \leq h_t~\forall t \in [T], \\
                                                   G_{t}(y_t, x_t) \leq g_t~\forall t \in [T]
                                                 \end{array} \!\!\!\! \right.\right\}.
\end{equation}

Note that the information map of the trajectory $X_{0:T}$ given a single observation $Y_{k}$ is given by
\begin{align}
\mathcal{I}_{X_{0:T}|Y_k}(y_k) &= \{ x_{0:T}~|~y_k \in P_{Y_k|X_k}(x_k)~\}, \\
                               &= \{ x_{0:T}~|~G_{k}(y_k, x_k) \leq g_k \}. \label{eq:kf_info1}
\end{align}
Also, the prior uncertainty set for $X_{0:T}$ is given by
\begin{equation}
\label{eq:kf_info2}
U_{X_{0:T}} = \{ x_{0:T}~|~H_{0}(x_0) \leq h_0~\}.
\end{equation}
From~\eqref{eq:kf_info1},~\eqref{eq:kf_info2}, and~\eqref{eq:kf_posti} we see that the posteriori uncertainty map in~\eqref{eq:kf_posti} can be written as an intersection of all the information maps and the prior uncertainty set:
\begin{multline}
P_{X_{0:T}|Y_{1:T}}(y_{1:T}) = U_{X_{0:T}|X_0}\cap \mathcal{I}_{X_{0:T}|Y_1}(y_1)\cap \mathcal{I}_{X_{0:T}|Y_2}(y_2)\cap \\
\ldots \cap\mathcal{I}_{X_{0:T}|Y_T}(y_T). \nonumber
\end{multline}
This affirms the triangulation principle proposed over trajectory space by Steven LaValle in~\cite{laValle_2012_sensing_filtering}.

\bibliographystyle{ieeetr}

\begin{thebibliography}{10}

\bibitem{talak19_Allerton_UV_BayNet}
R.~Talak, S.~Karaman, and E.~Modiano, ``A theory of uncertainty variables for
  state estimation and inference,'' in {\em Proc. {A}llerton}, Sep. 2019.

\bibitem{Hald_2007_ProbHistory1713Till1935}
A.~Hald, {\em A History of Parametric Statistical Inference from Bernoulli to
  Fisher, 1713-1935}.
\newblock Springer-Verlag, New York, 1~ed., 2007.

\bibitem{Stigler_1986_StatHistoryTill1900}
S.~M. Stigler, {\em The History of Statistics: The Measurement of Uncertainty
  Before 1900}.
\newblock Harvard University Press, Cambridge, 1~ed., 1986.

\bibitem{Kolmogorov_1933_ProbTheory}
A.~Kolmogorov, {\em Foundations of the Theory of Probability}.
\newblock 1933.

\bibitem{1978_Kingman_Exchangeability}
J.~F.~C. Kingman, ``Uses of exchangeability,'' {\em The Annals of Probability},
  vol.~6, no.~2, pp.~183--197, 1978.

\bibitem{1985_Aldous_Exchangeability}
D.~J. Aldous, ``Exchangeability and related topics,'' in {\em {\'E}cole
  d'{\'E}t{\'e} de Probabilit{\'e}s de Saint-Flour XIII --- 1983}, (Berlin,
  Heidelberg), pp.~1--198, Springer Berlin Heidelberg, 1985.

\bibitem{thurn_prob_robotics}
S.~Thrun, W.~Burgard, and D.~Fox, {\em Probabilistic Robotics}.
\newblock The MIT Press, 2005.

\bibitem{christian_robert_ml}
C.~Robert, {\em The Bayesian Choice}.
\newblock Springer-Verlag, New York, 2~ed., 2007.

\bibitem{bishop_ml}
C.~Bishop, {\em Pattern Recognition and Machine Learning}.
\newblock Springer-Verlag, New York, 1~ed., 2006.

\bibitem{2004_Complexity_MAP_BayNet}
J.~D. Park and A.~Darwiche, ``Complexity results and approximation strategies
  for map explanations,'' {\em J. Artif. Int. Res.}, vol.~21, pp.~101--133,
  Feb. 2004.

\bibitem{pmlr-v99-tosh19a}
C.~Tosh and S.~Dasgupta, ``The relative complexity of maximum likelihood
  estimation, map estimation, and sampling,'' {\em Machine Learning Research},
  vol.~99, pp.~2993--3035, Jun. 2019.

\bibitem{2001_minka_EP}
T.~P. Minka, ``Expectation propagation for approximate bayesian inference,'' in
  {\em Proc. Uncertainty in Artificial Intelligence}, pp.~362--369, 2001.

\bibitem{2002_particle_filtering}
M.~S. {Arulampalam}, S.~{Maskell}, N.~{Gordon}, and T.~{Clapp}, ``A tutorial on
  particle filters for online nonlinear/non-gaussian bayesian tracking,'' {\em
  IEEE Transactions on Signal Processing}, vol.~50, pp.~174--188, Feb 2002.

\bibitem{Andrieu2003_mcmc}
C.~Andrieu, N.~de~Freitas, A.~Doucet, and M.~I. Jordan, ``An introduction to
  mcmc for machine learning,'' {\em Machine Learning}, vol.~50, pp.~5--43, Jan.
  2003.

\bibitem{Blei2017_variational_inference}
D.~M. Blei, A.~Kucukelbir, and J.~D. McAuliffe, ``Variational inference: A
  review for statisticians,'' {\em Journal of the American Statistical
  Association}, vol.~112, no.~518, pp.~859--877, 2017.

\bibitem{koller_ProbGraphModels}
D.~Koller and N.~Friedman, {\em Probabilistic Graphical Models: Principles and
  Techniques}.
\newblock The MIT Press, 2009.

\bibitem{jordan_GraphModels}
M.~I. Jordan, ``Graphical models,'' {\em Statistical Science}, vol.~19,
  pp.~140--155, Jul. 2004.

\bibitem{Barefoot_2017_StateEstimation_Robotics}
T.~D. Barfoot, {\em State Estimation for Robotics}.
\newblock Cambridge University Press, Jul. 2017.

\bibitem{2015_Franco_SetEstiAndControl}
F.~Blanchini and S.~Miani, {\em Set-Theoretic Methods in Control}.
\newblock Birkh\"{a}user Basel, 2~ed., 2015.

\bibitem{1971_Ber_Thesis}
D.~Bertsekas, {\em Control of uncertain systems with a set-membership
  description of the uncertainity}.
\newblock {Ph.D.} Dissertation, MIT, 1971.

\bibitem{1971_BerRhodes_ReachabilityTubes}
D.~P. Bertsekas and I.~B. Rhodes, ``On the minimax reachability of target sets
  and target tubes,'' {\em Automatica}, vol.~7, pp.~233--247, Mar. 1971.

\bibitem{1989_deller_set_mem_identification_dsp}
J.~R. {Deller}, ``Set membership identification in digital signal processing,''
  {\em {IEEE} {ASSP} {M}agazine}, vol.~6, pp.~4--20, Oct. 1989.

\bibitem{1993_combettes_foundations_set_estimation}
P.~L. {Combettes}, ``The foundations of set theoretic estimation,'' {\em Proc.
  {IEEE}}, vol.~81, pp.~182--208, Feb. 1993.

\bibitem{2002_Ros_EllipsoidalCalculus}
L.~{Ros}, A.~{Sabater}, and F.~{Thomas}, ``An ellipsoidal calculus based on
  propagation and fusion,'' {\em IEEE Trans. Syst., Man, Cybern. B, Cybern.},
  vol.~32, pp.~430--442, Aug. 2002.

\bibitem{1997_Book_Ellipsoidal_Calculus}
A.~Kurzhanski and I.~Valyi, {\em Ellipsoidal Calculus for Estimation and
  Control}.
\newblock Birkh\"{a}user Basel, 1~ed., 1997.

\bibitem{2011_DBerm_RO_theory_n_applications}
D.~Bertsimas, D.~B. Brown, and C.~Caramanis, ``Theory and applications of
  robust optimization,'' {\em SIAM Review}, vol.~53, no.~3, pp.~464--501, 2011.

\bibitem{BenTal_RO_Book}
A.~Ben-Tal, L.~E. Ghaoui, and A.~Nemirovski, {\em Robust Optimization}.
\newblock Princeton University Press, 2009.

\bibitem{2011_DBerm_RO_high_dim}
C.~Bandi and D.~Bertsimas, ``Tractable stochastic analysis in high dimensions
  via robust optimization,'' {\em SIAM Review}, vol.~134, pp.~23--70, Aug.
  2012.

\bibitem{laValle_2012_sensing_filtering}
S.~M. LaValle, {\em Sensing and Filtering: A Fresh Perspective Based on
  Preimages and Information Spaces}, vol.~1.
\newblock Foundations and Trrends in Robotics, Feb. 2012.

\bibitem{1981_Whittle_Risk_Sensitive_LQGC}
P.~Whittle, ``Risk-sensitive linear/quadratic/gaussian control,'' {\em Advances
  in Applied Probability}, vol.~13, no.~4, pp.~764--777, 1981.

\bibitem{2019_Vasileios_OutlierRejection}
H.~Yang, P.~Antonante, V.~Tzoumas, and L.~Carlone, ``Graduated non-convexity
  for robust spatial perception: From non-minimal solvers to global outlier
  rejection,'' {\em arXiv e-prints arXiv:1909.08605}, Sep. 2019.

\bibitem{2002tac_risk_sensitive_filtering}
R.~K. {Boel}, M.~R. {James}, and I.~R. {Petersen}, ``Robustness and
  risk-sensitive filtering,'' {\em {IEEE} Transactions on Automatic Control},
  vol.~47, pp.~451--461, Mar. 2002.

\bibitem{1997tac_risk_sensitive_filtering_smoothing}
S.~{Dey} and J.~B. {Moore}, ``Risk-sensitive filtering and smoothing via
  reference probability methods,'' {\em {IEEE} Transactions on Automatic
  Control}, vol.~42, pp.~1587--1591, Nov. 1997.

\end{thebibliography}

\end{document}